\documentclass[11pt]{article}
\usepackage{coling2020}
\usepackage{times}
\usepackage{url}
\usepackage{latexsym}
\usepackage{multirow}
\usepackage{colortbl}
\usepackage{booktabs}
\usepackage{amssymb}
\usepackage{amsmath}
\usepackage{amsthm}
\usepackage{graphicx}
\usepackage{makecell}
\usepackage{threeparttable}
\usepackage{CJKutf8}
\usepackage{subcaption}
\usepackage{lscape}
\usepackage{fancyhdr}

\usepackage{listings} 
\usepackage{xcolor}
\colorlet{punct}{red!60!black}
\definecolor{background}{HTML}{EEEEEE}
\definecolor{delim}{RGB}{20,105,176}
\colorlet{numb}{magenta!60!black}

\lstdefinelanguage{json}{
    basicstyle=\small\ttfamily,
    numbers=left,
    numberstyle=\scriptsize,
    stepnumber=1,
    numbersep=6pt,
    showstringspaces=false,
    breaklines=true,
    frame=lines,
    backgroundcolor=\color{background},
    literate=
     *{0}{{{\color{numb}0}}}{1}
      {1}{{{\color{numb}1}}}{1}
      {2}{{{\color{numb}2}}}{1}
      {3}{{{\color{numb}3}}}{1}
      {4}{{{\color{numb}4}}}{1}
      {5}{{{\color{numb}5}}}{1}
      {6}{{{\color{numb}6}}}{1}
      {7}{{{\color{numb}7}}}{1}
      {8}{{{\color{numb}8}}}{1}
      {9}{{{\color{numb}9}}}{1}
      {:}{{{\color{punct}{:}}}}{1}
      {,}{{{\color{punct}{,}}}}{1}
      {\{}{{{\color{delim}{\{}}}}{1}
      {\}}{{{\color{delim}{\}}}}}{1}
      {[}{{{\color{delim}{[}}}}{1}
      {]}{{{\color{delim}{]}}}}{1},
}

\colingfinalcopy 

\newcommand*{\img}[1]{%
    \raisebox{-.5\baselineskip}{%
        \includegraphics[
        height=30pt,
        keepaspectratio,
        ]{#1}%
    }%
}
\definecolor{color_text}{RGB}{35, 70, 150}
\definecolor{color_flint}{RGB}{50, 150, 240}

\title{\img{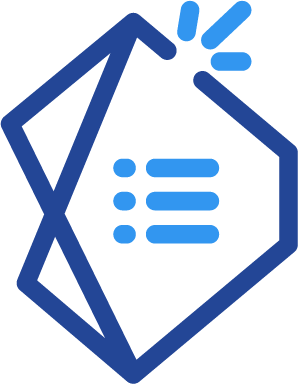} \textsf{\textcolor{color_text}{Text}\textcolor{color_flint}{Flint}}: Unified Multilingual Robustness Evaluation Toolkit for Natural Language Processing}

\author{Tao Gui$^{*}$, Xiao Wang\thanks{{} {} Tao Gui and Xiao Wang contributed equally to this work  and are co-first authors.}, Qi Zhang\thanks{{} {} Corresponding Author}, Qin Liu, Yicheng Zou, Xin Zhou, Rui Zheng, \\
 \bf Chong Zhang, Qinzhuo Wu, Jiacheng Ye, Zexiong Pang, Yongxin Zhang,  \\ 
 \bf Zhengyan Li, Ruotian Ma, Zichu Fei, Ruijian Cai, Jun Zhao, Xingwu Hu,  \\
 \bf Zhiheng Yan, Yiding Tan, Yuan Hu, Qiyuan Bian, Zhihua Liu, Bolin Zhu,\\ 
 \bf  Shan Qin,  Xiaoyu Xing, Jinlan Fu, Yue Zhang, Minlong Peng, \\ 
 \bf  Xiaoqing Zheng,  Yaqian Zhou, Zhongyu Wei, Xipeng Qiu and Xuanjing Huang \\
  School of Computer Science, Fudan University \\
  {\tt \{tgui16, xiao\_wang20, qz\}@fudan.edu.cn } }

\date{}

\begin{document}
\maketitle

\thispagestyle{fancy} 
\renewcommand{\headrulewidth}{0pt} 
\renewcommand{\footrulewidth}{0pt} 

\pagestyle{fancy} 
\fancyhead{\empty} 

\begin{abstract}
Various robustness evaluation methodologies from different perspectives have been proposed for different natural language processing (NLP) tasks. These methods have often focused on either universal or task-specific generalization capabilities. In this work, we propose a multilingual robustness evaluation platform for NLP tasks (\textsf{TextFlint}\footnote{\texttt{http://textflint.io/}}) that incorporates universal text transformation, task-specific transformation, adversarial attack, subpopulation, and their combinations to provide comprehensive robustness analysis. \textsf{TextFlint} enables practitioners to automatically evaluate their models from all aspects or to customize their evaluations as desired with just a few lines of code. To guarantee user acceptability, all the text transformations are linguistically based, and we provide a human evaluation for each one. \textsf{TextFlint} generates complete analytical reports as well as targeted augmented data to address the shortcomings of the model's robustness. To validate \textsf{TextFlint}'s utility, we performed large-scale empirical evaluations (over 67,000 evaluations) on state-of-the-art deep learning models, classic supervised methods, and real-world systems. Almost all models showed significant performance degradation, including a decline of more than 50\% of BERT's prediction accuracy on tasks such as aspect-level sentiment classification, named entity recognition, and  natural language inference. Therefore, we call for the robustness to be included in the model evaluation, so as to promote the healthy development of NLP technology.




\end{abstract}

\section{Introduction}
The recent breakthroughs in deep learning theory and technology provide strong support for the wide application of NLP technology, such as question answering systems \cite{seo2016bidirectional}, information extraction \cite{zeng2014relation}, and machine translation \cite{hassan2018achieving}. A large number of models have emerged, of which the performances surpass that of humans \cite{Lan2020ALBERT,Clark2020ELECTRA} when the training and test data are independent and identically distributed (i.i.d.). However, the repeated evaluation of models on a hold-out test set can yield overly optimistic estimates of the model performance \cite{dwork2015reusable}. The goal of building NLP systems is not merely to obtain high scores on the test datasets, but to generalize to new examples in the wild. However, recent research had reported that highly accurate deep neural networks (DNN) can be vulnerable to carefully crafted adversarial examples \cite{li2020bert}, distribution shift \cite{miller2020effect}, data transformation \cite{xing2020tasty}, and shortcut learning \cite{geirhos2020shortcut}. Using hold-out datasets that are often not comprehensive tends to result in trained models that contain the same biases as the training data \cite{rajpurkar2018know}, which makes it difficult to determine where the model defects are and how to fix them \cite{ribeiro-etal-2020-beyond}. 

Recently, researchers have begun to explore ways to detect robustness prior to model deployment. Approaches to textual robustness evaluation focus on making slight modifications to the input that maintain the original meaning but result in a different prediction. These approaches can be roughly divided into three categories: (1) adversarial attacks based on heuristic rules or language models that modify characters and substitute words \cite{morris2020textattack,zeng2020openattack}; (2) text transformations, task-agnostic \cite{ribeiro-etal-2020-beyond}, or task-specific \cite{xing2020tasty} testing methodologies that create challenge datasets based on specific natural language capabilities; (3) subpopulations that aggregate metrics with particular slices of interest \cite{wu2019errudite}. Using the continual evaluation paradigm rather than testing a static artifact, a model can continuously be evaluated in light of new information about its limitations. However, these methods have often focused on either universal or task-specific generalization capabilities, for which it is difficult to make a comprehensive robustness evaluation. We argue that the current robustness evaluations have the following three challenges:

\begin{enumerate}
    \item \textbf{Integrity.} When examining the robustness of a model, practitioners often hope that their evaluation is comprehensive and has verified the model's robustness from as many aspects as possible. However, previous work has often focused on universal or task-specific generalization capabilities. On one hand, universal generalization evaluations, like perturbations \cite{ribeiro-etal-2020-beyond} and subpopulations \cite{wu2019errudite}, have difficulty finding the core defects of different tasks (Section \ref{sec:task}). On the other hand, task-specific transformations may be invalid for use on other tasks. For customized needs (e.g., the combination of reversing sentiment and changing named entities), practitioners must try how to make different evaluation tools compatible.
    
    \item \textbf{Acceptability.} Only when newly transformed texts conforms to human language can the evaluation process obtain a credible robustness result. The uncontrollability of the words generated by a neural language model, incompatibility caused by template filling, and instability of heuristic rules in choosing words often make the generated sentences linguistically unacceptable to humans, which means the robustness evaluation will not be persuasive.
    
    \item \textbf{Analyzability.} Users require not only prediction accuracy on new datasets, but also relevant analyses based on these results. An analysis report should be able to accurately explain where a model's shortcomings lie, such as the problems with lexical rules or syntactic rules. Existing work has provided very little information regarding model performance characteristics, intended use cases, potential pitfalls, or other information to help practitioners evaluate the robustness of their models. This highlights the need for detailed documentation to accompany trained deep learning models, including metrics that capture bias, fairness and failure considerations \cite{10.1145/3287560.3287596}.
\end{enumerate}

In response to these challenges, here, we introduce \textsf{TextFlint}, a unified, multilingual, analyzable robustness evaluation toolkit for NLP. The challenges described above can be addressed in the \textbf{Customize} $\Rightarrow$ \textbf{Produce} $\Rightarrow$ \textbf{Analyze} workflow. We summarize this workflow as follows:

\begin{enumerate}
    \item \textbf{Customize.} \textsf{TextFlint} offers 20 general transformations and 60 task-specific transformations, as well as thousands of their combinations, which cover all aspects of text transformations to enable comprehensive evaluation of the robustness of a model (Section \ref{ling_trans}). \textsf{TextFlint} supports evaluations in multiple languages, currently English and Chinese, with other languages under development. In addition, \textsf{TextFlint} also incorporates adversarial attack and subpopulation. Based on the integrity of the text transformations, \textsf{TextFlint} automatically analyzes the deficiencies of a model with respect to its lexics, syntax, and semantics, or performs a customized analysis based on the needs of the user.  
    
    \item \textbf{Produce.} \textsf{TextFlint} provides 6,903 new evaluation datasets generated by the transformation of 24 classic datasets for 12 tasks. Users can directly download these datasets for robustness evaluation. For those who need comprehensive evaluation, \textsf{TextFlint} supports the generation of all the transformed texts and corresponding labels within one command, the automatic evaluation on the model, and the production of comprehensive analysis report. For those customized needs, users can modify the \texttt{Config} file and type a few lines of code to achieve a specific evaluation (Section \ref{framework}).

    \item \textbf{Analyze.} After scoring all of the existing transformation methods with respect to their plausibility and grammaticality by human evaluation, we use these results as a basis for assigning a confidence score for each evaluation result (Section \ref{human_eva}). Based on the evaluation results, \textsf{TextFlint} provides a standard analysis report with respect to a model's lexics, syntax, and semantic. All the evaluation results can be displayed via visualization and tabulation to help users gain a quick and accurate grasp of the shortcomings of a model. In addition, \textsf{TextFlint} generates a large number of targeted data to augment the evaluated model, based on the the defects identified in the analysis report, and provides patches for the model defects.
\end{enumerate}

\textsf{TextFlint} is easy to use for robustness analysis. To demonstrate the benefits of its process to practitioners, we outline how users with different needs can use \textsf{TextFlint} to evaluate their NLP models (Section \ref{workflow}). (1) Users who want to comprehensively evaluate a model's robustness can rely on predefined testbenches or generated datasets for direct evaluation. We explain how to use \texttt{FlintModel} automatically to evaluate model robustness from all aspects of text transformations (Section \ref{input_layer}). (2) Users who want to customize their evaluations for specific tasks can construct their own testbenches with a few lines of code using the \texttt{Config} available in \textsf{TextFlint}. (3) Users who want to improve model robustness can accurately identify the shortcomings of their model with reference to the analysis report (Section \ref{report_layer}), then use \textsf{TextFlint} to augment the training data for adversarial training(Section \ref{generation_layer}).

We tested 95 the state-of-the-art models and classic systems on 6,903 transformation datasets for a total of over 67,000 evaluations, and found almost all models showed significant performance degradation, including a decline of more than 50\% of BERT’s prediction accuracy on tasks such as aspect-level sentiment classification, named entity recognition, and natural language inference. It means that most experimental models are almost unusable in real scenarios, and the robustness needs to be improved.

\section{\textsf{TextFlint} Framework} \label{framework}
\textsf{TextFlint} provides comprehensive robustness evaluation functions, i.e., transformation, subpopulation and adversarial attack. For ordinary users, \textsf{TextFlint} provides reliable default config to generate comprehensive robustness evaluation data, with little learning cost. At the same time, \textsf{TextFlint} has strong flexibility and supports providing customized config files. \textsf{TextFlint} can automatically analyze the target model’s deficiencies and generate a visual report that can be used to inspire model improvement. Finally, \textsf{TextFlint} enables practitioners to improve their model by generating adversarial samples which can be used for adversarial training.

In this section, we introduce the design philosophy and modular architecture of \textsf{TextFlint}. In the following, workflow and usage for various requirements are provided.
\subsection{Design and Architecture}
\begin{figure}[t]
\centering
  \includegraphics[width=6.2in]{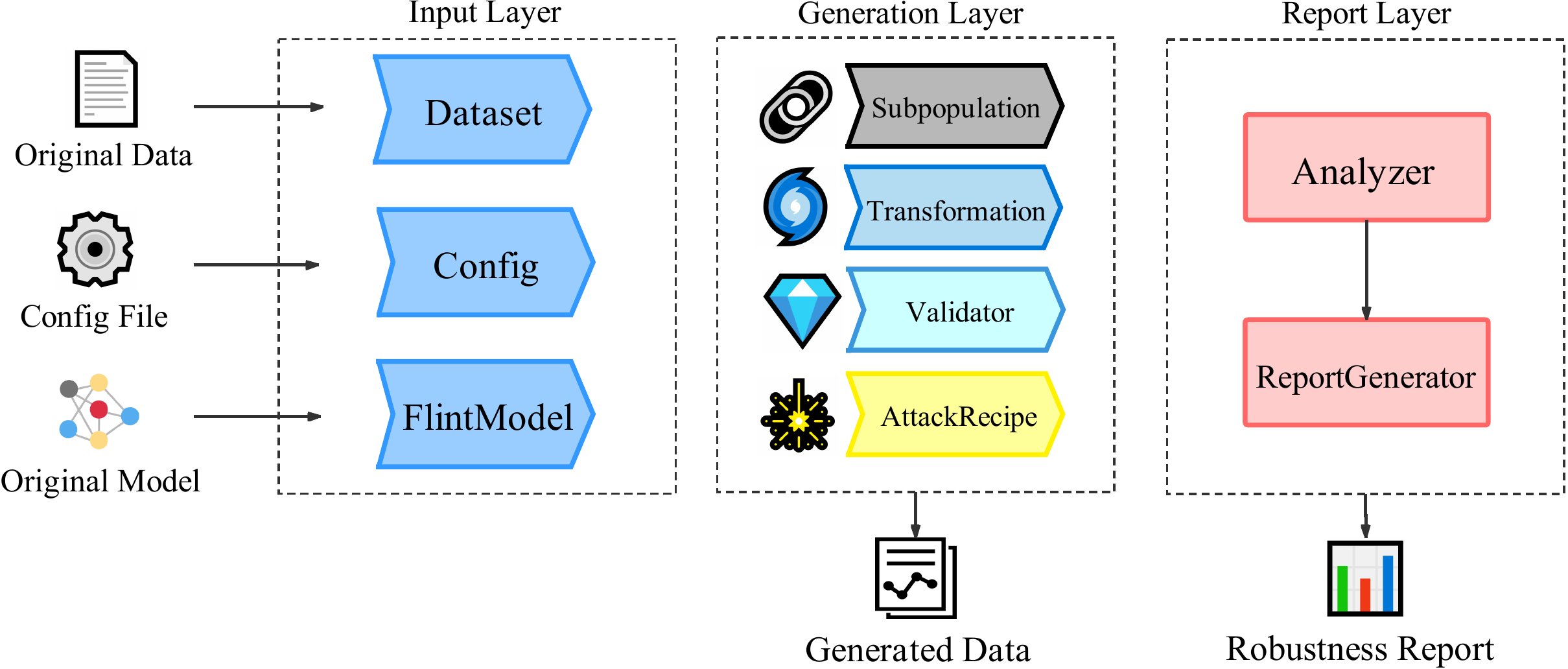}
  \caption{Architecture of \textsf{TextFlint}. Input Layer receives the original datasets, config files and target models as input, which are represented as \texttt{Dataset}, \texttt{Config} and \texttt{FlintModel} separately. Generation Layer consists of three parallel modules, where \texttt{Subpopulation} generates a subset of input dataset, \texttt{Transformation} augments datasets, and \texttt{AttackRecipe} interacts with the target model. Report Layer analyzes test results and provides users with robustness report.}
 \label{fig:Architecture}
\end{figure}
Figure \ref{fig:Architecture} shows the architecture of \textsf{TextFlint}. To apply \textsf{TextFlint} to various NLP tasks, its architecture is designed to be highly modular, easy to configure, and extensible.  \textsf{TextFlint} can be organized into three main components according to its workflow, i.e., Input Layer, Generation Layer, Reporter Layer, respectively. We will introduce each of the three components in more detail.

\subsubsection{Input Layer} \label{input_layer}
To apply robustness verification, Input Layer prepares necessary information, including the original dataset, config file, and target model.

\paragraph{Sample}
A common problem is that the input format of different models is highly different, making it very difficult to load and utilize data. It is therefore highly desirable to unify data structure for each task. \texttt{Sample} solves this problem by decomposing various NLP task data into underlying \texttt{Field}s, which cover all basic input types. \texttt{Sample} provides common linguistic functions, including tokenization, part-of-speech tagging and dependency parsing, which are implemented based on Spacy \cite{ines_montani_2021_4593273}. Moreover, we break down the arbitrary text transformation method into some atomic operations inside \texttt{Sample}, backed with clean and consistent implementations. Such design enables us to easily implement various transformations while reusing functions that are shared across transformations. 

\paragraph{Dataset}
\texttt{Dataset} contains samples and provides efficient and handy operation interfaces for samples. \texttt{Dataset} supports loading, verification, and saving data in JSON or CSV format for various NLP tasks. In addition, \textsf{TextFlint} integrates HuggingFace's NLP libraries \cite{wolf-etal-2020-transformers} which enable practitioners to download public datasets directly.

\paragraph{FlintModel}
\texttt{FlintModel} is a necessary input to apply adversarial attack or generate robustness report. \textsf{TextFlint} has great extensibility and allows practitioners to customize target model with whichever deep learning framework. Practitioners just need to wrap their own models through \texttt{FlintModel} and implement the corresponding interfaces. 

\paragraph{Config}
It is vital for the toolkit to be flexible enough to allow practitioners to configure the workflow, while providing appropriate abstractions to alleviate the concerns of practitioners who overly focus on the low-level implementation. \textsf{TextFlint} enables practitioners to provide a customized config file to specify certain types of \texttt{Tranformation}, \texttt{Subpopulation}, \texttt{AttackRecipe} or their combinations, as well as their related parameters information. Of course, \textsf{TextFlint} provides reliable default parameters, which reduces the threshold for use.

\subsubsection{Generation Layer} \label{generation_layer}
After Input Layer completes the required input loading, the interaction between \textsf{TextFlint} and the user is complete. Generation Layer aims to apply data generation function which includes \texttt{Transformation}, \texttt{Subpopulation} and \texttt{AttackRecipe} to each sample. To improve memory utilization, Generation Layer dynamically creates \texttt{Transformation}, \texttt{SubPopulation}, and \texttt{AttackRecipe} instances according to the parameters of the \texttt{Config} instance.

\paragraph{Transformation}
Based on the atomic operations provided by Sample, it is easy to implement an arbitrary text transformation while ensuring the correctness of the transformation. Thanks to the highly modular design of \textsf{TextFlint}, Transformation can be flexibly applied to samples for different tasks.It is worth noting that the procedure of \texttt{Transformation} does not need to query the target model, which means it is a completely decoupled process with the target model prediction.

In order to verify the robustness comprehensively, \textsf{TextFlint} offers 20 universal transformations and 60 task-specific transformations, covering 12 NLP tasks. According to the granularity of the transformations, the transformations can be categorized into sentence level, word level and character level. Sentence-level transformations includes \textbf{BackTranslation}, \textbf{Twitter}, \textbf{InsertAdv}, etc. Word-level transformations include \textbf{SwapSyn-WordNet}, \textbf{Contraction}, \textbf{MLMSuggestion}, etc. Character level deformation includes \textbf{KeyBoard}, \textbf{Ocr}, \textbf{Typos}, etc. Due to limited space, refer to Section~\ref{ling_trans} for specific information. 

\paragraph{AttackRecipe}
\texttt{AttackRecipe} aims to find a perturbation of an input text satisfies the attack's goal to fool the given \texttt{FlintModel}. In contrast to \texttt{Transformation}, \texttt{AttackRecipe} requires the prediction scores of the target model. Once \texttt{Dataset} and \texttt{FlintModel} instances are provided by Input Layer, \textsf{TextFlint} would apply \texttt{AttackRecipe} to each sample. \textsf{TextFlint} provides 16 easy-to-use adversarial attack recipes which are implemented based on TextAttack \cite{morris2020textattack}.

\paragraph{Validator}
Are all generated samples correct and retain the same semantics as the original samples, instead of being completely unrecognizable by humans? It is crucial to verify the quality of samples generated by \texttt{Transformation} and \texttt{AttackRecipe}. \textsf{TextFlint} provides several metrics to calculate confidence, including (1) language model perplexity calculated based on the GPT2 model \cite{radford2019language}; (2) word replacement ratio in the generated text compared with the original text; (3)The edit distance between original text and generated text; (4) Semantic similarity calculated based on Universal Sentence Encoder \cite{cer-etal-2018-universal}; (5) BLEU score \cite{papineni-etal-2002-bleu}.

\paragraph{Subpopulation}
\texttt{Subpopulation} is to identify the specific part of dataset on which the target model performs poorly. To retrieve a subset that meets the configuration, \texttt{Subpopulation} divides the dataset through sorting samples by certain attributes.
\textsf{TextFlint} provides 4 general \texttt{Subpopulation} configurations, including text length, language model performance, phrase matching, and gender bias, which work for most NLP tasks.
Take the configuration of text length for example, \texttt{Subpopulation} retrieves the subset of the top 20\% or bottom 20\% in length.

\subsubsection{Report Layer} \label{report_layer}
In Generation Layer, \textsf{TextFlint} can generate three types of adversarial samples and verify the robustness of the target model.
Based on the results from Generation Layer, Report Layer aims to provide users with a standard analysis report from lexics, syntax, and semantic levels.
The running process of Report Layer can be regarded as a pipeline from \texttt{Analyzer} to \texttt{ReportGenerator}.

\paragraph{Analyzer}

The \texttt{Analyzer} is designed to analyze the robustness of the target model from three perspectives: (1) robustness against multi-granularity transformed data and adversarial attacks; (2) gender bias and location bias; (3) subpopulation division. 
For the shortcomings of the target model,  \texttt{Analyzer} can also look for potential performance improvement directions.

\paragraph{ReportGenerator}

\begin{figure}[t]
\centering
  \includegraphics[width=6.3in]{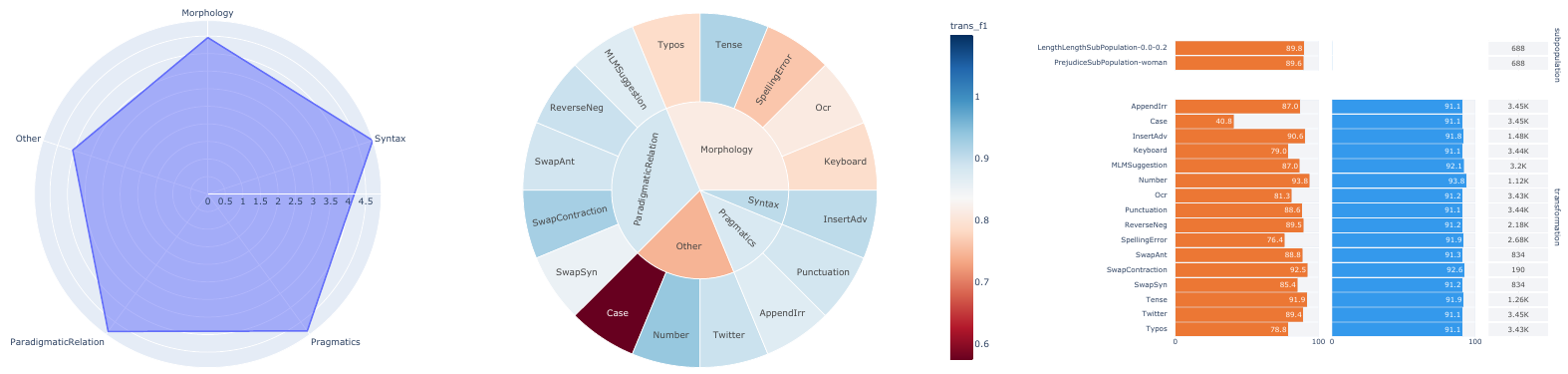}
  \caption{Robustness reports of BERT base model on CONLL 2003 dataset. The first one, namely the radar report, provides an overview of the linguistic ability of the target model. The middle chart gives an intuitive result on each transformation categorized by linguistics. The last bar-chart reveals the details of model performance towards every single generation method.
  } \label{fig:bert_conll}
\end{figure}

According to the analysis provided by \texttt{Analyzer}, \texttt{ReportGenerator} can visualize and chart the performance changes of the target model under different transformations.
\texttt{ReportGenerator} conveys the analysis results to users in PDF or \LaTeX \ format, which makes it easy to save and share. 
\texttt{ReportGenerator} also provides users with a concise and elegant API to display their results and reduce the cost of analysis in a large amount of experimental data.
We believe that a clear and reasonable analysis report will inspire users.
Take BERT base\cite{devlin2019bert} model on CONLL 2003\cite{sang2003introduction} dataset as example, its robustness report is displayed in Figure \ref{fig:bert_conll}. 

\subsection{Usage}

Using \textsf{TextFlint} to verify the robustness of a specific model is as simple as running the following command:

\begin{lstlisting}[language=sh]
$ textflint --dataset input_file --config config.json
\end{lstlisting}
where \texttt{input\_file} is the input file of csv or json format, \texttt{config.json} is a configuration file with generation and target model options. 

Complex functions can be implemented by a simple modification on \texttt{config.json}, such as executing the pipeline of transformations and assigning the parameters of each transformation method. Take the configuration for TextCNN~\cite{kim2014convolutional} model on SA (sentiment analysis) task as example:

\begin{lstlisting}[language=json]
{
  "task": "SA",
  "out_dir": "./DATA/",
  "flint_model": "./textcnn_model.py",
  "trans_methods": [
    "Ocr",
    ["InsertAdv", "SwapNamedEnt"],   
    ...
  ],
  "trans_config": {
    "Ocr": {"trans_p": 0.3},
    ...
  },
...
}
\end{lstlisting}

\begin{itemize}
\item  \texttt{task} is the name of target task. \textsf{TextFlint} supports 12 types of tasks. For task names, please refer to the official website description document \url{https://github.com/textflint/textflint}.

\item  \texttt{out\_dir} is the directory where each of the generated sample and its corresponding original sample are saved.

\item  \texttt{flint\_model} is the python file path that saves the instance of FlintModel.

\item \texttt{trans\_methods} is used to specify the transformation method. For example, \texttt{"Ocr"} denotes the universal transformation \textbf{\emph{Ocr}}, and \texttt{["InsertAdv", "SwapNamedEnt"]} denotes a pipeline of task-specific transformations, namely \textbf{\emph{InsertAdv}} and \textbf{\emph{SwapNamedEnt}}.

\item \texttt{trans\_config} configures the parameters for the transformation methods. The default parameter is also a good choice. 
\end{itemize}

Moreover, it also supports the configuration of subpopulation and adversarial attack. For more details about parameter configuration, please move to \url{https://github.com/textflint/textflint}.

\begin{figure}[t]
\begin{center}
\includegraphics[width=10.5cm]{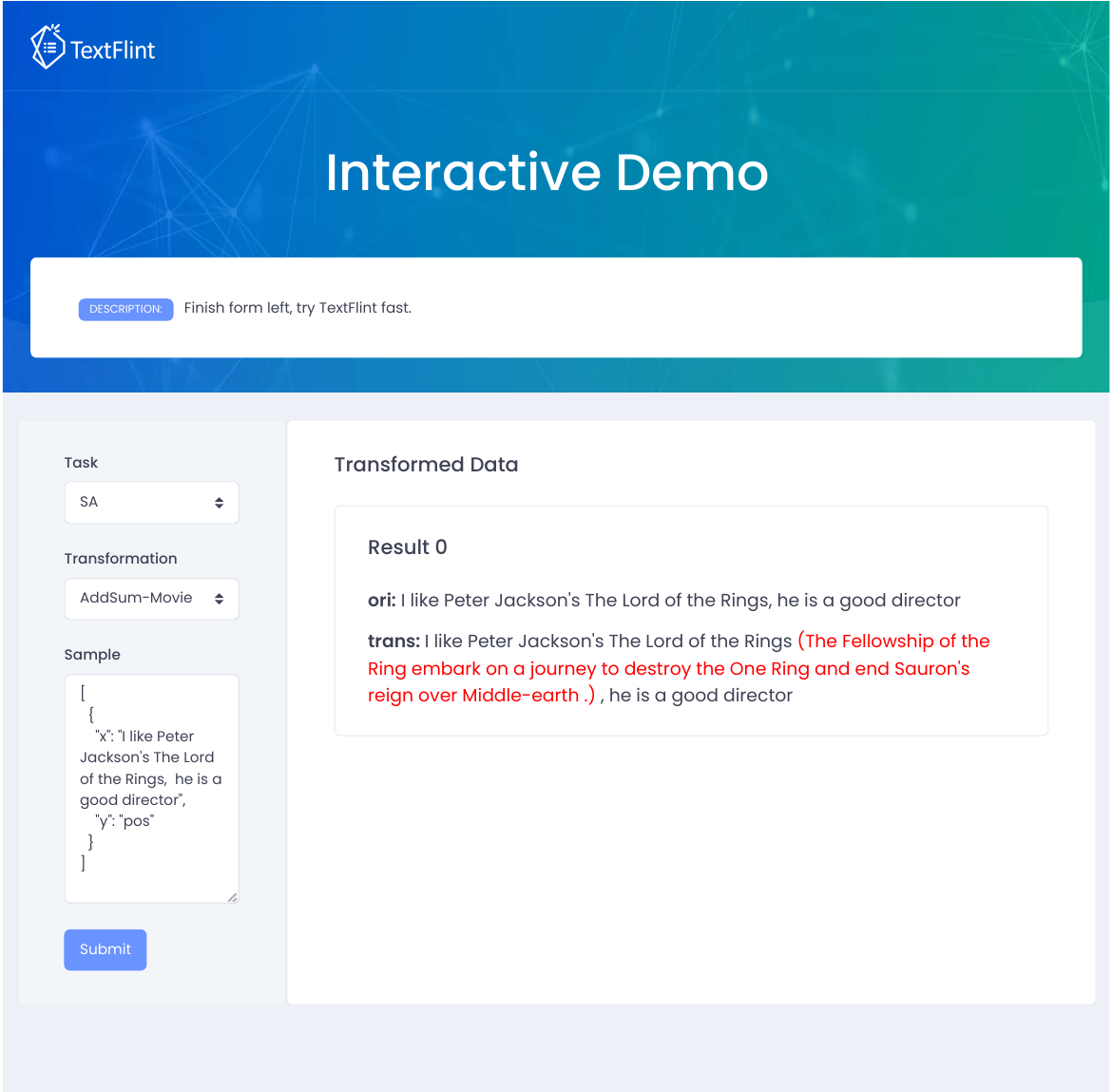}
\end{center}
  \caption{Screenshot of \textsf{TextFlint}’s web interface running \textbf{\emph{AddSum-Movie}} transformation for SA task.}
\label{fig:result}
\end{figure}

Based on the design of decoupling sample generation and model verification, \textsf{TextFlint} can be used inside another NLP project with just a few lines of code.
\begin{lstlisting}[language=Python]
from textflint import Engine

data_path = 'input_file'
config = 'config.json'
engine = Engine()
engine.run(data_path, config)
\end{lstlisting}

\textsf{TextFlint} is also available for use through our web demo, displayed in Figure~\ref{fig:result}, which is available at
\url{https://www.textflint.io/demo}.

\subsection{Workflow} \label{workflow}

\begin{figure}[t]
\centering
  \includegraphics[width=6.3in]{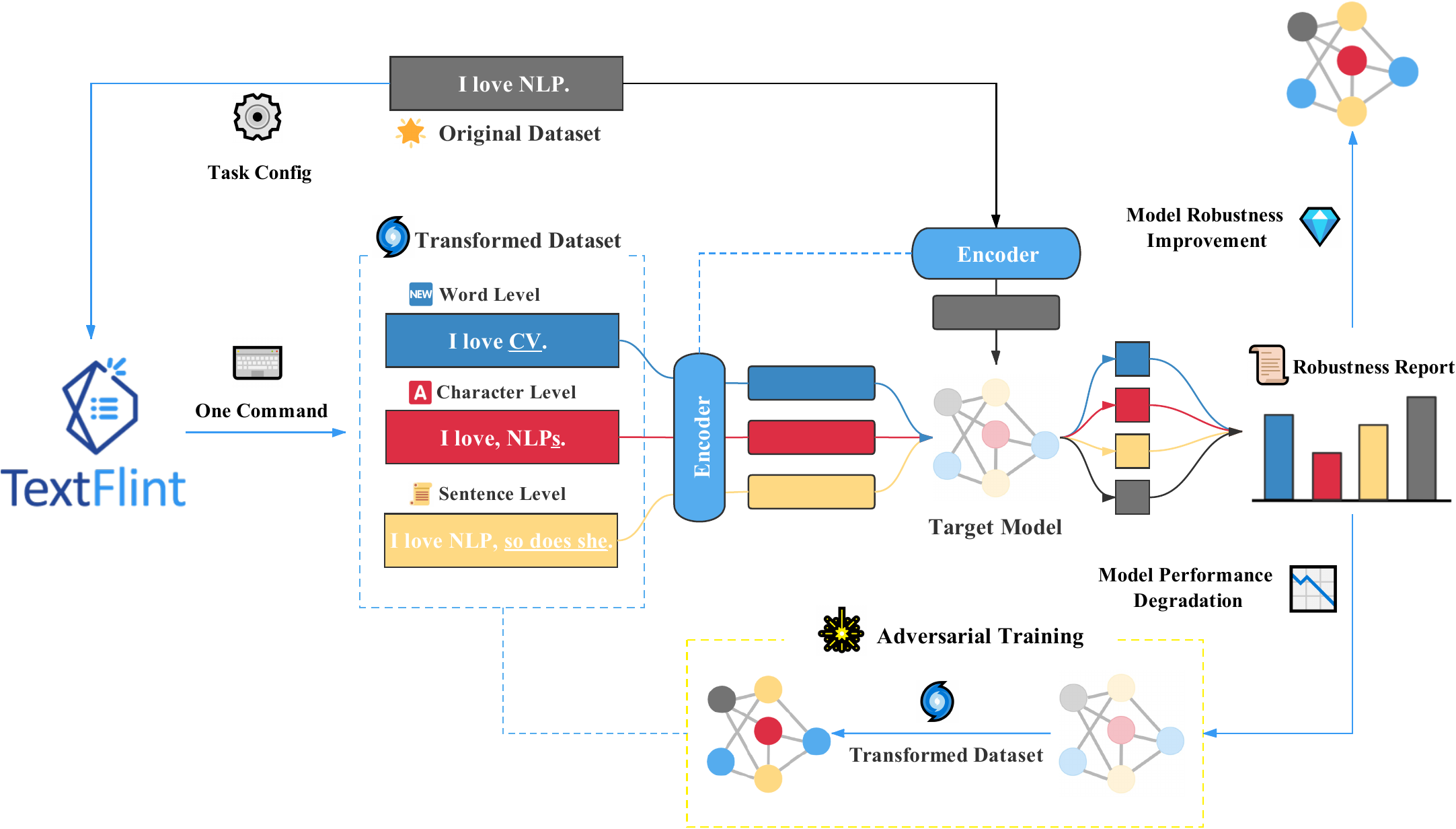}
  \caption{Workflow of \textsf{TextFlint}. Original dataset is transformed in \textsf{TextFlint} by multi-granularity transformations, which is specified by task config. The original and transformed datasets are then applied to target models to evaluate model robustness on multiple transformations. Results will finally be reported in a visualized form, and transformed dataset could further be used as adversarial training samples of target models. 
  } \label{fig:workflow}
\end{figure}

The general workflow of \textsf{TextFlint} is displayed in Figure \ref{fig:workflow}. With correspondence to Figure \ref{fig:Architecture}, evaluation of target models could be devided into three steps.
For input preparation, the original dataset for testing, which is to be loaded by \texttt{Dataset}, should be firstly formatted as a series of JSON objects. \textsf{TextFlint} configuration is specified by \texttt{Config}. Target models are also loaded as \texttt{FlintModel}s. 
Then in adversarial sample generation, multi-perspective transformations (as \texttt{Transformation}), including subpopulation division (as \texttt{Subpopulation}), are performed on \texttt{Dataset} to generate transformed samples. Besides, to ensure semantic and grammatical correctness of transformed samples, \texttt{Validator} calculates confidence of each sample to filter out unacceptable samples. 
Lastly, \texttt{Analyzer} collects evaluation results and \texttt{ReportGenerator} automatically generates a comprehensive report of model robustness. 
Additionally, users could feed train dataset into \textsf{TextFlint} to obtain substantial amount of transformed samples, which could be used to do adversarial training on target models.

Due its user-friendly design philosophy, \textsf{TextFlint} shows its practicality in real application.
As mentioned in Section 1, we summarize three occasions in which users would found challenging in model robustness evaluation. In those occasions, \textsf{TextFlint} is proven to be helpful due to its comprehensive features and customization ability.

\paragraph{General Evaluation} For users who want to evaluate robustness of NLP models in a general way, \textsf{TextFlint} supports generating massive and comprehensive transformed samples within one command. By default, \textsf{TextFlint} performs all single transformations on original dataset to form corresponding transformed datasets, and the performance of target models is tested on these datasets. As a feedback of model robustness, the results of target model performance change on each of the transformed datasets and their corresponding original datasets are reported in a clear form. The evaluation report provides a comparative view of model performance on datasets before and after certain types of transformation, which supports model weakness analysis and guides particular improvement. 

\paragraph{Customized Evaluation} For users who want to test model performance on specific aspects, they demand a customized transformed dataset of certain transformations or their combinations. In \textsf{TextFlint}, this could be achieved by modifying \texttt{Config}, which determines the configuration of \textsf{TextFlint} in generation. \texttt{Config} specifies the transformations being performed on the given dataset, and it could be modified manually or generated automatically. 
Moreover, by modifying the configuration, users could decide to generate multiple transformed samples on each original data sample, validate samples by semantics, preprocess samples with certain processors, etc. 

\paragraph{Target Model Improvement} For users who want to improve robustness of target models, they may work hard to inspect the weakness of model with less alternative support. To tackle the issue, we believe a diagnostic report revealing the influence of comprehensive aspects on model performance would provide concrete suggestions on model improvement. By using \textsf{TextFlint} and applying transformed dataset to target models, the transformations corresponding to significant performance decline in evaluation report will provide improvement guidance of target models.
Moreover, \textsf{TextFlint} supports adversarial training on target models with large-scale transformed dataset, and the change of performance will also be reported to display performance gain due to adversarial training.

To summarize, the ease-to-use framework satisfies the needs of model robustness evaluation by providing multi-aspect transformations and supporting automatic analysis. Moreover, the proposed transformation schemes in \textsf{TextFlint} are ensured to be linguistic-conformed and human-accepted, which liberates users from contemplating and implementing their own transformation schemes. In the next section, the linguistic basis of transformations included in \textsf{TextFlint} will be concisely discussed. 

\section{Linguistically based Transformations} \label{ling_trans}
\begin{figure}[t]
\centering
  \includegraphics[width=6.3in]{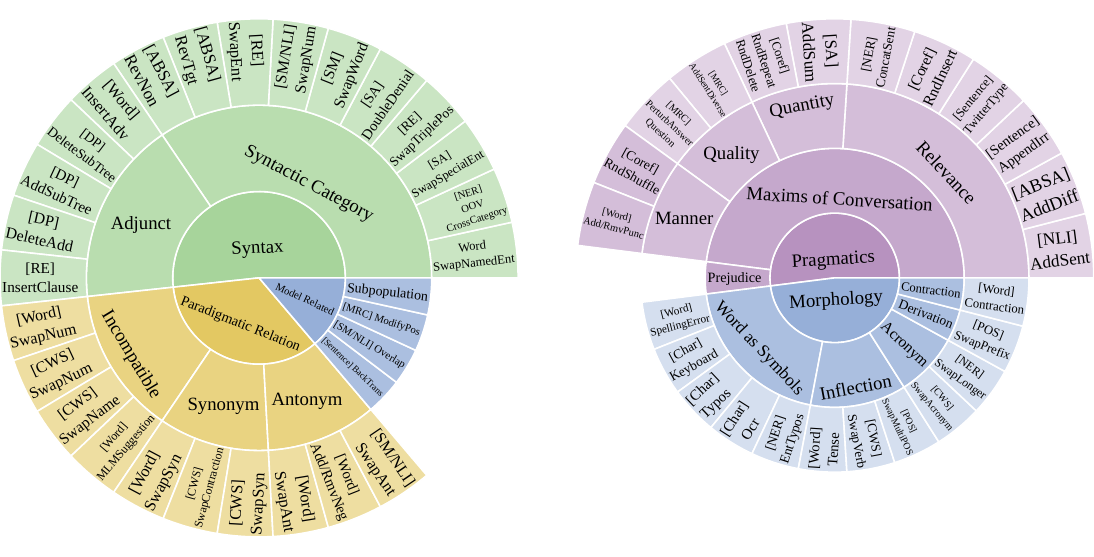}
  \caption{Overview of transformations through the lens of linguistics.
  } \label{fig:ling-trans}
\end{figure}

We attempt to increase the variety of text transformations to a large extent while maintaining the acceptability of transformed texts. For this purpose, we turn to linguistics for inspiration and guidance (Figure~\ref{fig:ling-trans}), which is to be discussed at length in the following sections with \textbf{bold} for universal transformations and \textbf{\emph{bold italic}} for task-specific ones.

\subsection{Morphology}
With word-level transformation being the first step, morphology sheds light on our design from the very beginning. Morphology is the study of how words are formed and interrelated. It analyzes the structure of words and parts of words, e.g., stems, prefixes, and suffixes, to name a few. This section discusses the transformations with respect to different aspects of morphology.

\subsubsection{Derivation}
Morphological derivation is the process of forming a new word from an existing word, often by adding a prefix or suffix, such as \emph{ab-} or \emph{-ly}. For example, \emph{abnormal} and \emph{normally} both derive from the root word \emph{normal}.

Conversion, also called ``zero derivation'' or ``null derivation,'' is worth noting as well. It is a type of word formation involving the creation of a word from an existing word without any change in form, namely, derivation using only zero. For example, the noun \emph{green} is derived ultimately from the adjective \emph{green}. That is to say, some words, which can be derived with zero, carry several different parts of speech.

\paragraph{\emph{SwapPrefix}}
Swapping the prefix of one word usually keeps its part of speech.\footnote{Few prefixes can change the POS of the root word: \textbf{en-}, \textbf{de-}, \textbf{be-}, \textbf{a-}, and \textbf{out-}. For instance, danger (n.) $\rightarrow$ \textbf{en-}danger(v.); grade (n.)$\rightarrow$ \textbf{de-}grade(v.); friend (n.) $\rightarrow$ \textbf{be-}friend (v.); sleep (n.) $\rightarrow$ \textbf{a-}sleep(adv.); rank (n.) $\rightarrow$ \textbf{out-}rank(v.). These prefixes are not considered for text transformation.} 
For instance, \emph{``This is a \textbf{pre-}fixed string''} might be transformed into \emph{``This is a \textbf{trans-}fixed string''} or \emph{``This is an \textbf{af-}fixed string.''}
The POS tags of the test sentence is supposed to remain the same, since it is merely changed in one single word without converting its part of speech. \emph{\textbf{SwapPrefix}} is especially applicable to the POS tagging task in NLP.

\paragraph{\emph{SwapMultiPOS}}
It is implied by the phenomenon of conversion that some words hold multiple parts of speech. That is to say, these multi-part-of-speech words might confuse the language models in terms of POS tagging. Accordingly, we replace nouns, verbs, adjectives, or adverbs with words holding multiple parts of speech, e.g., ``\emph{There is an apple on the desk}'' is transformed into ``\emph{There is an imponderable on the desk}'' by swapping the noun \emph{apple} into \emph{imponderable}, which can be a noun or an adjective. Although the transformed sentence is not as accessible as the original, anyone with even the slightest knowledge of English would be able to tell the right part of speech of \emph{imponderable} that fits the context without understanding its meaning. Since the transformation of \emph{\textbf{SwapMultiPOS}} alters the semantic meaning of sentences, it is, again, only applicable for the POS tagging task.

\subsubsection{Inflection}
Morphological inflection generally tells the tense, number, or person of one word. The word ``love'', for example, performs differently in sentences ``\emph{I love NLP},'' ``\emph{He love\textbf{-s} NLP},'' ``\emph{She love\textbf{-d} NLP},'' and ``\emph{They are love\textbf{-ing} NLP},'' where \emph{love\textbf{-s}} denotes that the subject of the verb \emph{love} is third person and singular and that the verb is in the present tense, while \emph{love\textbf{-d}} denotes the simple past tense and \emph{love\textbf{-ing}} for present progressive. Similarly, the transformation \textbf{Tense} changes the tense of verbs while maintaining the semantic meaning to a large extent, just as from ``\emph{He is studying NLP}'' to ``\emph{He has studied NLP}.''

Besides, reduplication is a special type of inflection in which the root or stem of a word or even the whole word is repeated exactly or with a slight change. As, for example, \emph{quack-quack} imitates the sound of a duck, \emph{fiddle-faddle} suggests something of inferior quality, and \emph{zigzag} suggests alternating movements. This phenomenon is more common in Chinese than in English, where most verbs with one character ``A'' can be reduplicated to express the same meaning in the form of ``A(?)A'', just as the verb \begin{CJK}{UTF8}{gbsn} ``看 (look)''\end{CJK} holds the same meaning with \begin{CJK}{UTF8}{gbsn} ``看看,''\end{CJK} \begin{CJK}{UTF8}{gbsn} ``看一看,''\end{CJK} \begin{CJK}{UTF8}{gbsn} ``看了看,''\end{CJK} and \begin{CJK}{UTF8}{gbsn} ``看了一看.''\end{CJK}
As a result, the accordingly implemented \emph{\textbf{SwapVerb}} is tailored especially for the task of Chinese word segmentation.

\subsubsection{Contraction}
A contraction is a word made by shortening and combining two words, such as \emph{can't} (can + not), \emph{you're} (you + are), and \emph{I've} (I + have), which is often leveraged in both speaking and writing. Contraction changes the form of words while leaving the semantic meaning unchanged. Likewise, the transformation \textbf{Contraction} replaces phrases like \emph{will not} and \emph{he has} with contracted forms, namely, \emph{won't} and \emph{he's}. With \textbf{Contraction} modifying neither the syntactic structure nor the semantic meaning of the original sentence, it fits all of the tasks in NLP, be it token- or sequence-level.

\subsubsection{Acronym}
An acronym is a shorter version of an existing word or phrase, usually using individual initial letters or syllables, as in \textbf{NATO} (North Atlantic Treaty Organization) or \textbf{App} (application). From the perspective of acronym, \emph{\textbf{SwapLonger}} detects the acronyms in one sentence and supersedes them with the full form, with \emph{NLP} changed into \emph{Natural Language Processing} and \emph{USTC} into \emph{University of Science and Technology of China}. Although \emph{\textbf{SwapLonger}} might be feasible for most NLP tasks, it is especially effective in evaluating the robustness of models for named entity recognition (NER) in that it precisely modifies those named entities to be recognized. Similarly, \emph{\textbf{SwapAcronym}} is tailored for Chinese word segmentation in a reverse way that the acronym like \begin{CJK}{UTF8}{gbsn} ``中国 (China)''\end{CJK} is turned into its full form \begin{CJK}{UTF8}{gbsn} ``中华人民共和国 (People's Republic of China)''\end{CJK} to confuse the segmentation.

\subsubsection{Word as Symbols}
As a tool for communication, language is often written as symbols, as it is also regarded as a symbol system. Thus, words have the ``form-meaning duality.'' From time to time, a typographical error happens while writing or typing, which means the form of a word is destructed while the meaning stays. Humans often make little effort to understand words with typographical errors; however, such words might be totally destructive for deep learning models.

To imitate this common condition in the daily use of language, \textbf{SpellingError} and \textbf{Typos} both bring slight errors to words, while being implemented in different ways. The former replaces a word with its typical error form (\emph{definitely} $\rightarrow$ \emph{difinately}), and the latter randomly inserts, deletes, swaps or replaces a single letter within one word (\emph{Ireland} $\rightarrow$ \emph{Irland}). Nearly the same with \textbf{Typos}, \emph{\textbf{EntTypos}} works for NER and swaps only named entities with misspelled ones (\emph{Shanghai} $\rightarrow$ \emph{Shenghai}). \textbf{Keyboard} turn to the way how people type words and change tokens into mistaken ones with errors caused by the use of keyboard, like \emph{word} $\rightarrow$ \emph{worf} and \emph{ambiguous} $\rightarrow$ \emph{amviguius}. Besides, it is worth noting that some texts are generated from pictures by optical character recognition (OCR); we also take into consideration the related errors. With \emph{like} being recognized as \emph{l1ke} or \emph{cat} as \emph{ca+}, human readers take no effort understanding these mistaken words, while it is worth inspecting how language models react toward this situation.

\subsection{Paradigmatic Relation}
A paradigmatic relation describes the type of semantic relations between words that can be substituted with another word in the same category, which contains synonymy, hyponymy, antonymy, etc. As in the sentence ``\emph{I read the ( ) you wrote two years ago,}'' the bracket can be filled with \emph{book}, \emph{novel}, \emph{dictionary}, or \emph{letter}. The following sections discuss the specific relations leveraged in our transformations.

\subsubsection{Synonym}
A synonym is a word or phrase that means nearly the same as another word or phrase. For example, the words \emph{begin}, \emph{start}, \emph{commence}, and \emph{initiate} are all synonyms of one another. One synonym can be replaced by another in a sentence without changing its meaning. Correspondingly, \textbf{SwapSyn} (\textbf{Syn} short for synonym) switches tokens into their synonyms according to WordNet or word embedding. As for instance, ``\emph{He loves NLP}'' is transformed into ``\emph{He likes NLP}'' by simply replacing \emph{loves} with \emph{likes}.

\subsubsection{Antonym}
Antonymy describes the relation between a pair of words with opposite meanings. For example, \emph{mortal} : \emph{immortal}, \emph{dark} : \emph{light}, and \emph{early} : \emph{late} are all pairs of antonyms. Although the meaning of a sentence is altered after one word being replaced by its antonym, the syntax of the sentence remain unchanged.
As a result, \textbf{SwapAnt} and \textbf{Add/RmvNeg} are suitable for some NLP tasks, including but not limited to dependency parsing, POS tagging, and NER. The implementation of \textbf{SwapAnt} is similar with \textbf{SwapSyn}, while \textbf{Add/RmvNeg} performs differently. Transferring ``\emph{John lives in Ireland}'' into ``\emph{John doesn't live in Ireland},'' the overall meaning is reversed with a simple insertion of the negation \emph{doesn't}, while the syntactic structure is saved.

\subsubsection{Incompatible}
Incompatibility is the relation between two classes with no members in common. Two lexical items X and Y are incompatibles if ``$A\ is\  f(X)$'' entails ``$A\ is\ not\ f(Y)$'': ``\emph{I'm from Shanghai}'' entails ``\emph{I'm not from Beijing},'' where \emph{Shanghai} and \emph{Beijing} are a pair of incompatibles. Incompatibility makes up for the vacancy, where neither synonym nor antonym is applicable. The transformation \textbf{SwapNum}, which means swap number, shifts numbers into different ones, such as ``\emph{Tom has 3 sisters}'' into ``\emph{Tom has 2 sisters}.'' \textbf{\emph{SwapName}} is designed specially for Chinese word segmentation, which is not simply substituting people's names into random others but deliberately chosen ones of which the first character can form a phrase with the former character. To make it clear, \begin{CJK}{UTF8}{gbsn} ``我朝\textbf{小明}挥了挥手 (
I waved at Xiao Ming)''\end{CJK} might be changed into \begin{CJK}{UTF8}{gbsn} ``我朝\textbf{向明}挥了挥手 (
I waved at Xiang Ming),''\end{CJK} where the first character, \begin{CJK}{UTF8}{gbsn} ``\textbf{向},''\end{CJK} of the substitution forms a phrase with the character \begin{CJK}{UTF8}{gbsn} ``朝,''\end{CJK} resulting in \begin{CJK}{UTF8}{gbsn} ``朝向（towards）.''\end{CJK} Though the semantic meaning is slightly changed with the swap of the mentioned name, the result of segmentation is supposed to remain the same.

\subsection{Syntax}
The rules of syntax combine words into phrases and phrases into sentences, which also specify the correct word order for a language, grammatical relations of a sentence as well as other constraints that sentences must adhere to. In other words, syntax governs the structure of sentences from various aspects.

\subsubsection{Syntactic Category}
A family of expressions that can substitute for one another without loss of grammaticality is called a syntactic category, including both lexical category, namely the part of speech, and phrasal category. To illustrate, in ``\emph{I love NLP}'' and ``\emph{I love CV},'' \emph{NLP} and \emph{CV} belong to the same lexical category of noun (N); for ``\emph{He is running in the park}'' and ``\emph{He is running on the roof},'' \emph{in the park} and \emph{on the roof} are of the same phrasal category, namely, prepositional phrase (PP), which means these two phrases are interchangeable without altering the structure of the whole sentence.

Realizing this special component of a sentence makes room for various transformations at this level. \textbf{SwapNamedEnt}, \emph{\textbf{SwapSpecialEnt}}, and \emph{\textbf{SwapWord/Ent}} are clear as their names imply, the named entities in a sentence are swapped into others of the same type, which means the syntactic structure and the named entity tags remain constant. Similarly, \emph{\textbf{OOV}} and \emph{\textbf{CrossCategory}} are special to the NER task, where the substitutions are out of vocabulary or are from a different category. For instance, ``\emph{I love NLP}'' can be transformed into either ``\emph{I love NlP}'' (\emph{\textbf{OOV}}) or ``\emph{I love Shanghai}'' (\emph{\textbf{CrossCategory}}). \emph{\textbf{DoubleDenial}}, tailored for semantic analysis (SA), is able to preserve both syntactic and semantic attribute of the original sentence, such as ``\emph{I love NLP}'' and ``\emph{I don't hate NLP}.'' For aspect based semantic analysis (ABSA), \emph{\textbf{RevTgt}} (short for reverse target) and \emph{\textbf{RevNon}} (short for reverse non-target) generate sentences that reverse the original sentiment of the target aspect and the non-target aspect respectively. As in the sentence ``\emph{Tasty burgers, and crispy fries,}'' with the target aspect being \emph{burgers}, it might be transformed into ``\emph{Terrible burgers, but crispy fries}'' by \emph{\textbf{RevTgt}} or ``\emph{Tasty burgers, but soggy fries}'' by \emph{\textbf{RevNon}}.

Similarly, \textbf{MLMSuggestion} (\textbf{MLM} short for masked language model) generates new sentences where one syntactic category element of the original sentence is replaced by what is predicted by masked language models. With the original sentence ``\emph{This is a good history lesson}'' masked into ``\emph{This is a good (),}'' \textbf{MLMSuggestion} generates several predictions like \emph{story}, \emph{data set}, or \emph{for you}, of which the first two are retained to augment test data in that they are of the same syntactic category with \emph{history lesson}.

Besides commuting a single syntactic category element with brand new ones, the existing two elements within one sentence can be shifted with one another. \emph{\textbf{SwapTriplePos}} (\textbf{Pos} short for position) exchanges the position of two entities in one triple, which works only for the relation extraction (RE) task. For example, the sentence ``\emph{Heig, born in Shanghai, was graduated from Fudan University,}'' where subject and object are \emph{Heig} and \emph{Shanghai}, and the relation being birth, can be transformed into ``\emph{Born in Shanghai, Heig was graduated from Fudan University.}''

\subsubsection{Adjunct}
An adjunct is a structurally dispensable part of a sentence that, if removed or discarded, will not structurally affect the remainder of the sentence. As in the sentence ``\emph{I love NLP from bottom of my heart},'' the phrase \emph{from bottom of my heart} is an adjunct, which also means a modifier of the verb \emph{love}. Since the adjunct is structurally dispensable, it doesn't matter if any adjunct is removed or appended. As typical adjuncts, adverbs can be inserted before the verb of one sentence without considering the structure or semantics, \textbf{InsertAdv} is adaptable for most of the NLP tasks. Furthermore, \emph{\textbf{Delete/AddSubTree}} and \emph{\textbf{InsertClause}} change sentences by appending or removing adjuncts from the aspect of dependency parsing (DP), just as the sub tree/clause ``\emph{who was born in China}'' being inserted into the original sentence ``\emph{He loves NLP}'' and ends up with ``\emph{He, who was born in China, loves NLP}.''

\subsection{Pragmatics}
Pragmatics concerns how context affects meaning, e.g., how the sentence ``\emph{It’s cold in here}'' comes to be interpreted as  ``\emph{close the windows}'' in certain situations. Pragmatics explain how we are able to overcome ambiguity since meaning relies on the manner, place, time, etc. of an utterance.

\subsubsection{Maxims of Conversation}
The maxims of conversation were first discussed by the British philosopher H. Paul Grice and are sometimes called ``Gricean Maxims.'' These maxims describe how people achieve effective conversational communication in common social situations, including the maxim of quantity, quality, relevance, and manner. The maxim of quantity requires saying neither more nor less than the discourse requires. The maxim of quality suggests not telling lies or making unsupported claims. The maxim of relevance, just as the name implies, tells that the speakers should say what is relevant to the conversation. Last but not least, the maxim of manner values being perspicuous, namely, avoiding obscurity or ambiguity of expression and being brief and orderly.

Grice did not assume that all people should constantly follow these maxims. Instead, it is found to be interesting when these were not respected, which is going to be further illustrated in the following sections. For example, with Marry saying, ``\emph{I don't think the model proposed by this paper is robust},'' her listener, Peter, might reply that ``\emph{It's a lovely day, isn't it?}'' Peter is flouting the maxim of relevance in that the author of the paper being discussed is standing right behind Marry who doesn't even notice a bit. From time to time, people flout these maxims either on purpose or by chance, while listeners are still able to figure out the meaning that the speaker is trying to convey, inspired by which we design transformations to simulate the situations where the maxims of conversation are flouted. 

\emph{\textbf{AddSum}} (\emph{\textbf{Sum}} short for summary) and \emph{\textbf{RndRepeat/Delete}} (\emph{\textbf{Rnd}} short for random) imitates flouting the maxim of quantity in a way that providing more or less information than needed. \emph{\textbf{AddSum}} works for semantic analysis, which involves adding the summary of the mentioned movie or person, enriching the background information of the sentence however unnecessary. \emph{\textbf{RndRepeat/Delete}} 
proves effective for the paragraph-level task of coreference resolution, where the number of sentences makes room for repetition and deletion, providing inappropriate amount of information for the purpose of communication.

For the same reason, \emph{\textbf{RndShuffle}} is also made possible by coreference resolution in a way of going against the maxim of manner, which randomly shuffles sentences within one paragraph and messes up the logic chain of utterance and causes confusion. Another transformation that reflects the offence of manner maxim is \textbf{Add/RmvPunc} (short for remove punctuation), namely, adding extra punctuation or removing necessary ones to disturb target models.

Considering the offence of maxim of quality, \emph{\textbf{AddSentDiverse}} (\textbf{Sent} short for sentence) and \emph{\textbf{PerturbAnswer/Question}} for machine reading comprehension (MRC) tasks bring disturbance to either the texts based on which questions are to be answered or the formulation of questions.

Last but not least, the maxim of relevance is also often flouted by language users. Analogously, we inspect language models' performance upon the springing of irrelevant information with \textbf{AppendIrr} (\textbf{Irr} short for irrelevant), \textbf{TwitterType}, \emph{\textbf{RndInsert}}, and \emph{\textbf{ConcatSent}}. The first two transformations adapt to most NLP tasks in that they change texts without altering the semantic meaning or the original structure of sentences. Specifically, \textbf{TwitterType} changes plan texts into the style of Twitter posts, such as turning ``\emph{I love NLP}'' into ``\emph{@Smith I love NLP. https://github.com/textflint}.'' \emph{\textbf{RndInsert}} and \emph{\textbf{ConcatSent}} work for coreference resolution and NER respectively.

\subsubsection{Language and Prejudice}
Words of a language reflect individual or societal values~\cite{IntrotoLang}, which can be seen in phrases like ``\emph{masculine charm}'' and ``\emph{womanish tears}.'' Until recently, most people subconsciously assume a professor to be a man and a nurse to be a woman. Besides, users of any language might also relate countries or regions with certain values. It's clear that language reflects social bias toward gender and many other aspects, and also any social attitude, positive or negative.

For inspection of how language models take on the prejudice that resides in human language, \textbf{Prejudice} offers the exchange of mentions of either human or region, from mentions of male into female or from mentions of one region into another. To specify, ``\emph{Marry loves NLP and so does Ann}'' can be replaced by ``\emph{Tom loves NLP and so does Jack}'' by a simple setting of ``male,'' which means all the mentions of female names are altered into names of male. The settings of region are just alike.

\subsection{Model Related}
Besides how humans use language, we also take into consideration how deep learning models actually process language to design transformations accordingly. We examine the general patterns of language models and end up with the following transformations.

\textbf{BackTrans} (\textbf{Trans} short for translation) replaces test data with paraphrases by leveraging back translation, which is able to figure out whether or not the target models merely capture the literal features instead of semantic meaning. \emph{\textbf{ModifyPos}} (\textbf{\emph{Pos}} short for position), which works only for MRC, examines how sensitive the target model is to the positional feature of sentences by changing the relative position of the golden span in a passage. As for the task of natural language inference (NLI), the overlap of the premise and its hypothesis is an easily captured yet unexpected feature. To tackle this, \emph{\textbf{Overlap}} generates pairs of premise and hypothesis by overlapping these two sentences and making a slight difference on the hypothesis, just as premise being ``\emph{The judges heard the actors resigned}'' and its hypothesis being ``\emph{The judges heard the actors}.''

The aforementioned three transformations focus on examining the features learned by target models, and \textbf{Subpopulation} tackles the distribution of a dataset by singling out a subpopulation of the whole test set in accordance with certain rules. To be more specific, \textbf{LengthSubpopulation} retrieves subpopulation by the length of each text, and \textbf{LMSubpopulation} (\textbf{LM} short for language model) by the performance of the language model on certain test data, for both of which the top 20\% and bottom 20\% are available as options.


\subsection{Human Evaluation} \label{human_eva}
Only when transformed text conforms to the way how humans use language can the evaluation process obtain a credible robustness result. To verify the quality of the transformed text, we conducted human evaluation on the original and transformed texts under all of the above mentioned transformations. Specifically, we consider two metrics in human evaluation, i.e., plausibility and grammaticality.

\begin{itemize}
\item \textbf{Plausibility} \cite{lambert2010creating} measures whether the text is reasonable and written by native speakers. Sentences or documents that are natural, appropriate, logically correct, and meaningful in the context will receive a higher plausibility score. Texts that are logically or semantically inconsistent or contain inappropriate vocabulary will receive a lower plausibility score.
\item \textbf{Grammaticality} \cite{newmeyer1983grammatical} measures whether the text contains syntax errors. It refers to the conformity of the text to the rules defined by the specific grammar of a language.
\end{itemize}

For human evaluation, we used the text generated from both the universal and task-specific transformations to compare with the original text from all of the twelve NLP tasks.
We randomly sample 100 pairs of original and transformed texts for each transformation in each task, with a total of about 50,000 texts.
We invite three native speakers from Amazon Mechanical Turk to evaluate the plausibility and grammaticality of these texts and record the average score. For each metric, we ask the professionals to rate the texts on a 1 to 5 scale (5 for the best).
Due to the limitations of the layout, we select tasks based on four common NLP problems: text classification, sequence labeling, semantic matching, and semantic understanding. The human evaluation results of these tasks are shown in Table \ref{table_HE_UT} and Table \ref{table_HE_DT}, and the remaining results are available at \texttt{http://textflint.io}.

We have the following observations:
\begin{enumerate}
  \item The human evaluation score is consistent and reasonable on the original text of each task, which proves the stability and effectiveness of our human evaluation metrics. From table \ref{table_HE_UT}, we can see that the human evaluation scores of the original text are consistent within each task. For the Grammaticality metric, the scores for all four tasks are around 3.7. One possible explanation for this case is that the source datasets of these original texts are well organized and have no obvious grammatical errors. For the plausibility metric, ABSA scores the highest, ranging from 4 to 4.1, while MRC scores are the lowest, ranging from 3.3 to 3.5. ABSA data are about restaurant reviews, and a single topic leads clear logic. MRC data are long paragraphs on various topics, a large number of proper nouns and domain-specific knowledge, making it more difficult to judge the rationality of these texts.
  \item The transformed text generated by the universal transformations can be accepted by humans. As shown in Table \ref{table_HE_UT}, different transformation methods change the original text to different degrees and result in different human evaluation scores. Some transformations (e.g., \textbf{WordCase}, \textbf{AddPunc} change the case of text or add/delete punctuations. These transformations do not change the semantics of the text or affect the readability, so their human evaluation scores did not change much. Some transformations (e.g., \textbf{SwapSyn}, \textbf{SwapAnt}) replace several words in the original text with their synonyms or antonyms. These transformations are well developed and widely used, and they will slightly lower the evaluation scores. Some transformations (e.g., \textbf{Ocr}, \textbf{SpellingError}, and \textbf{Tense}) replace words in the text with wrong words or change the tense of verbs. These transformations actively add wrong information to the original text and cause the human evaluation score to decrease.  On the whole, the transformed text has achieved a competitive human evaluation performance compared with the original text in each task. This verifies that, when the text has pattern changes, minor spelling errors, and redundant noisy information, these transformed texts are still fluent and readable and therefore acceptable to humans.
  \item The transformed text generated by the task-specific transformations still conforms to human language habits, while task-specific transformations change the original text more than universal transformations. As shown in Table \ref{table_HE_DT}, we believe this is because these transformations are specific to each task, and they have a good attack effect on the original text, which leads to larger changes in the text. The \textbf{\emph{ConcatSent}} transformation in the NER task concatenates multiple original texts into one text. The transformed text has no grammar error, but the logic between different sentences is inconsistent. As a result, its Plausibility drops from 4.14 to 3.54 while Grammaticality remains the same. In the SA task, the movie and person vocab list contains common phrases, such as  ``go home'', and these transformations may contain grammar errors, resulting in varying degrees of Grammaticality decline. However, replacing the movie and person names has little effect on the rationality of the sentence, and the Plausibility remains unchanged. The evaluation performance of these transformations is still stable and acceptable. This proves, again, that the transformed texts conform to human language, and the robustness evaluation results with these transformed texts are also persuasive.
\end{enumerate}

\begin{landscape}
\begin{table*}[!t]
  \centering
  \caption{Human evaluation results for universal transformations. Ori. and Trans. represent the original text and the transformed text, respectively. These metrics are rated on a 1-5 scale (5 for the best).}\label{table_HE_UT}
  \resizebox{1.54\textwidth}{!}{%
    \begin{tabular}{l|cccccccccccccccc}
    \hline
    &\multicolumn{4}{c}{ABSA}& \multicolumn{4}{c}{POS} & \multicolumn{4}{c}{NLI}& \multicolumn{4}{c}{MRC}\\
    \cmidrule(r){2-5}\cmidrule(r){6-9}\cmidrule(r){10-13}\cmidrule(r){14-17}
     & \multicolumn{2}{c}{Plausibility}& \multicolumn{2}{c}{Grammaticality} & \multicolumn{2}{c}{Plausibility}& \multicolumn{2}{c}{Grammaticality} & \multicolumn{2}{c}{Plausibility}& \multicolumn{2}{c}{Grammaticality} & \multicolumn{2}{c}{Plausibility}& \multicolumn{2}{c}{Grammaticality}\\
    \cmidrule(r){2-3}\cmidrule(r){4-5}\cmidrule(r){6-7}\cmidrule(r){8-9}
    \cmidrule(r){10-11}\cmidrule(r){12-13}\cmidrule(r){14-15}\cmidrule(r){16-17}
    &Ori.&Trans.&Ori.&Trans.&Ori.&Trans.&Ori.&Trans.
    &Ori.&Trans.&Ori.&Trans.&Ori.&Trans.&Ori.&Trans.\\
    \hline
    \textbf{InsertAdv} &4.08 	&4.06 	&3.97 	&3.80 	&3.72	&3.55	&3.68	&3.59	&3.73	&3.72	&3.7	&3.58	&3.5	&3.41	&3.79	&3.73\\
    \textbf{AppendIrr}	&4.08 	&4.05 	&3.96 	&3.90 	&3.73	&3.66	&3.7	&3.64	&3.62	&3.62	&3.63	&3.67	&3.48	&3.29	&3.8	&3.74\\
    \textbf{BackTrans}	&——	&——	&——	&——	&——	&——	&——	&——	&3.79	&3.94	&3.69	&3.59	&——	&——	&——	&——\\
    \textbf{WordCase-lower} &4.03 	&4.10 	&3.99 	&3.76 	&3.69	&3.53	&3.68	&3.67	&3.74	&3.67	&3.65	&3.72	&3.36	&3.34	&3.79	&3.67\\
    \textbf{WordCase-title} &4.04 	&4.13 	&3.95 	&3.74 	&3.67	&3.6	&3.65	&3.66	&3.68	&3.52	&3.57	&3.37	&3.36	&3.34	&3.76	&3.75\\
    \textbf{WordCase-upper} &4.15 	&4.11 	&3.99 	&3.79 	&3.69	&3.54	&3.68	&3.68	&3.78	&3.83	&3.65	&3.76	&3.36	&3.2	&3.76	&3.53\\
    \textbf{Contraction}	&4.01 	&4.02 	&4.04 	&4.01 	&3.56	&3.56	&3.7	&3.76	&3.66	&3.65	&3.57	&3.47	&3.32	&3.42	&3.74	&3.77\\
    \textbf{SwapNamedEnt } &4.18 	&3.99 	&3.94 	&3.97 	&3.62	&3.61	&3.72	&3.85	&3.74	&3.67	&3.56	&3.58	&——	&——	&——	&——\\
    \textbf{Keyboard}	&3.97 	&4.04 	&3.99 	&3.30 	&3.73	&3.41	&3.7	&3.33	&3.68	&3.51	&3.74	&3.48	&3.37	&3.3	&3.78	&3.57\\
    \textbf{MLMSuggestion} &4.23 	&4.14 	&3.93 	&3.75 	&3.71	&3.7	&3.68	&3.63	&3.73	&3.68	&3.58	&3.55	&——	&——	&——	&——\\
    \textbf{SwapNum} &4.35 	&4.13 	&4.10 	&3.67 	&3.38	&3.56	&3.74	&3.72	&3.68	&3.49	&3.47	&3.63	&——	&——	&——	&——\\
    \textbf{Ocr}	&4.12 	&4.03 	&4.15 	&3.12 	&3.73	&3.52	&3.7	&3.54	&3.59	&3.56	&3.71	&3.21	&3.36	&3.3	&3.76	&3.41\\
    \textbf{AddPunc} &4.12 	&4.10 	&4.10 	&3.65 	&3.73	&3.69	&3.7	&3.67	&3.57	&3.78	&3.54	&3.57	&3.36	&3.25	&3.75	&3.74\\
    \textbf{ReverseNeg}	&——	&——	&——	&——	&3.64	&3.62	&3.65	&3.82	&——	&——	&——	&——	&——	&——	&——	&——\\
    \textbf{SpellingError}	&4.07 	&3.80 	&3.80 	&3.30 	&3.7	&3.44	&3.7	&3.5	&3.55	&3.65	&3.54	&3.61	&3.36	&3.29	&3.76	&3.63\\
    \textbf{Tense}	&3.99 	&4.04 	&3.99 	&3.75 	&3.69	&3.55	&3.71	&3.79	&3.76	&3.6	&3.73	&3.65	&3.36	&3.16	&3.76	&3.56\\
    \textbf{TwitterType } &4.13 	&4.03 	&3.91 	&3.87 	&3.73	&3.54	&3.7	&3.73	&3.75	&3.6	&3.7	&3.56	&3.39	&3.24	&3.77	&3.67\\
    \textbf{Typos} &4.11 	&4.07 	&3.89 	&3.17 	&3.73	&3.46	&3.7	&3.42	&3.57	&3.69	&3.59	&3.43	&3.36	&3.31	&3.76	&3.54\\
    \textbf{SwapSyn-WordEmbedding}	&4.02 	&4.11 	&3.99 	&3.94 	&3.7	&3.52	&3.68	&3.82	&3.79	&3.46	&3.68	&3.67	&3.36	&3.51	&3.76	&3.7\\
    \textbf{SwapSyn-WordNet}	&——	&——	&——	&——	&3.68	&3.45	&3.72	&3.75	&——	&——	&——	&——	&——	&——	&——	&——\\
    \textbf{SwapAnt-WordNet}	&4.09 	&4.03 	&4.01 	&3.77 	&3.71	&3.67	&3.68	&3.72	&3.69	&3.82	&3.58	&3.6	&3.36	&3.23	&3.76	&3.64\\

    \hline
    \hline
    \end{tabular}
  }

  \end{table*}
\end{landscape}

\begin{table}[t]
  \caption{Human evaluation results for task-specific transformation. Ori. and Trans. represent the original text and the transformed text, respectively. These metrics are rated on a 1-5 scale (5 for the best).}\label{table_HE_DT}
\begin{minipage}[t]{0.53\linewidth}
\centering
\subcaption{SA}
  \resizebox{1\textwidth}{!}{%
\begin{tabular}{l|cccc}
\hline
& \multicolumn{2}{c}{Plausibility}& \multicolumn{2}{c}{Grammaticality}  \\
    \cmidrule(r){2-3}\cmidrule(r){4-5}
    &Ori.&Trans.&Ori.&Trans. \\
 \hline
    \textbf{\emph{DoubleDenial}} &3.26 	&3.37 	&3.59 	&3.49 \\
    \textbf{\emph{AddSum-Person}} &3.39 	&3.32 	&3.76 	&3.59 \\
    \textbf{\emph{AddSum-Movie}}&3.26 	&3.34 	&3.61 	&3.58 \\
    \textbf{\emph{SwapSpecialEnt-Person}} &3.37 	&3.14 	&3.75 	&3.73 \\
    \textbf{\emph{SwapSpecialEnt-Movie}}&3.17 	&3.28 	&3.70 	&3.49 \\

 \hline
\end{tabular}
}
\end{minipage}\begin{minipage}[t]{0.45\linewidth}
\centering
\subcaption{NER}
  \resizebox{1\textwidth}{!}{%
\begin{tabular}{l|cccc}
\hline
& \multicolumn{2}{c}{Plausibility}& \multicolumn{2}{c}{Grammaticality}  \\
    \cmidrule(r){2-3}\cmidrule(r){4-5}
    &Ori.&Trans.&Ori.&Trans. \\
 \hline
\textbf{\emph{OOV}}	&3.69	&3.76	&3.54	&3.48\\
\textbf{\emph{SwapLonger}}	&3.73	&3.66	&3.77	&3.54\\
\textbf{\emph{EntTypos}}	&3.57	&3.5	&3.59	&3.54\\
\textbf{\emph{CrossCategory}}	&3.48	&3.44	&3.41	&3.32\\
\textbf{\emph{ConcatSent}}	&4.14	&3.54	&3.84	&3.81\\

 \hline
\end{tabular}
}
\end{minipage}

\hfill

\begin{minipage}[t]{0.5\linewidth}
\centering
\subcaption{SM}
  \resizebox{1\textwidth}{!}{%
\begin{tabular}{l|cccc}
\hline
& \multicolumn{2}{c}{Plausibility}& \multicolumn{2}{c}{Grammaticality}  \\
    \cmidrule(r){2-3}\cmidrule(r){4-5}
    &Ori.&Trans.&Ori.&Trans. \\
 \hline
    \textbf{\emph{SwapWord}} &3.08	&3.08	&3.98	&3.92\\
    \textbf{\emph{SwapNum}}	&3.14	&3.21	&3.87 	&3.86\\
    \textbf{\emph{Overlap}}	&—	&3.33	&—	&4.11\\
 \hline
\end{tabular}
}
\end{minipage}
\begin{minipage}[t]{0.46\linewidth}
\centering
\subcaption{RE}
  \resizebox{1\textwidth}{!}{%
\begin{tabular}{l|cccc}
\hline
& \multicolumn{2}{c}{Plausibility}& \multicolumn{2}{c}{Grammaticality}  \\
    \cmidrule(r){2-3}\cmidrule(r){4-5}
    &Ori.&Trans.&Ori.&Trans. \\
 \hline
    \textbf{\emph{SwapEnt-MultiType}}	&3.59	&3.36	&3.97	&3.94\\
    \textbf{\emph{SwapEnt-LowFreq}}	&3.34	&3.56	&3.94	&4.05\\
    \textbf{\emph{InsertClause}}	&3.37	&3.4	&3.89	&3.95\\
    \textbf{\emph{SwapEnt-AgeSwap}}	&3.29	&3.52	&3.85	&4.07\\
    \textbf{\emph{SwapTriplePos-BirthSwap}} &3.52	&3.53	&3.91	&3.86\\
    \textbf{\emph{SwapTriplePos-EmployeeSwap}}	&3.39	&3.43	&3.88	&3.86\\
 \hline
\end{tabular}
}
\end{minipage}
\end{table}

\section{Evaluations with \textsf{TextFlint}}

\subsection{Task-Specific Transformation}\label{sec:task}

We conduct comprehensive experiments on 12 NLP tasks with a variety of models to present robustness evaluation results. For each task, we select at least one classic dataset  and perform linguistically based transformations on the test set to generate new test samples. For all existing models, we use the authors' official implementations, which are evaluated on original test samples and new samples to show the change of model performance. Here, we demonstrate five representative tasks that cover different languages and domains. For each task, we present results of more than 10 different models under a variety of task-specific transformations. 

{\bf Aspect-Based Sentiment Analysis (ABSA)} is a typical text classification task that aims to identify fine-grained sentiment polarity toward a specific aspect associated with a given target. In this work, we conduct experiments on SemEval 2014 Laptop and Restaurant Reviews \cite{pontiki2014semeval}, one of the most popular ABSA datasets, to test robustness of different lines of systems, including SOTA neural architectures. We follow Xu et al. \shortcite{xu2019bert} to remove instances with conflicting polarity and use the same train-dev split strategy. In the experiment, we adopt {\em Accuracy} and {\em Macro-F1} as the metrics to evaluate system performances, which are widely used in previous works \cite{fan2018multi,xing2020tasty}. 

Table \ref{tab:task_absa} shows the results of 10 different models on the SemEval 2014 Restaurant dataset. Based on the original test set, we choose 847 test instances with obvious opinion words to produce new samples. Finally, we obtain 847, 582, and 847 test instances in the transformation settings of \textbf{\emph{RevTgt}}, \textbf{\emph{RevNon}}, and \textbf{\emph{AddDiff}}, respectively. From the table, we can see that both the accuracy and macro-F1 scores of all models on the original restaurant test set are very high, achieving nearly 86\% on accuracy and 65\% on macro-F1. Nevertheless, they drop significantly on all three new test sets. \textbf{\emph{RevTgt}} leads to the most performance drop, as it requires the model to pay precise attention to the target sentiment words \cite{xing2020tasty}. \textbf{\emph{AddDiff}} causes drastic performance degradation among non-BERT models, indicating that these models lack the ability to distinguish relevant aspects from non-target aspects.

\begin{table*}[!t]
\small
\centering
\caption{Accuracy and F1 score on the SemEval 2014 Restaurant dataset.
}
\resizebox{\textwidth}{!}{
\begin{tabular}{lcccccc}
\toprule
\multirow{2}{*}{Model}
& \multicolumn{2}{c}{\textbf{\emph{RevTgt}} (Ori. $\rightarrow$ Trans.)}
& \multicolumn{2}{c}{\textbf{\emph{RevNon}} (Ori. $\rightarrow$ Trans.)}
& \multicolumn{2}{c}{\textbf{\emph{AddDiff}} (Ori. $\rightarrow$ Trans.)}
\\
& Accuracy & Macro-F1
& Accuracy & Macro-F1
& Accuracy & Macro-F1
\\
\midrule
\multicolumn{7}{l}{\textbf{\textit{Restaurant Dataset}}}\\
LSTM \cite{hochreiter1997long} & \textcolor[rgb]{0.0,0,0}{84.42 $\rightarrow$ 19.30} & \textcolor[rgb]{0.0,0,0}{55.75 $\rightarrow$ 19.88} & \textcolor[rgb]{0.0,0,0}{85.91 $\rightarrow$ 73.42} & \textcolor[rgb]{0.0,0,0}{55.02 $\rightarrow$ 44.69} & \textcolor[rgb]{0.0,0,0}{84.42 $\rightarrow$ 44.63} & \textcolor[rgb]{0.0,0,0}{55.75 $\rightarrow$ 33.24}
\\
TD-LSTM \cite{tang2016effective}& \textcolor[rgb]{0.0,0,0}{86.42 $\rightarrow$ 22.42 } & \textcolor[rgb]{0.0,0,0}{61.92 $\rightarrow$ 22.28} & \textcolor[rgb]{0.0,0,0}{87.29 $\rightarrow$ 79.58} & \textcolor[rgb]{0.0,0,0}{60.70 $\rightarrow$ 53.35} & \textcolor[rgb]{0.0,0,0}{84.42 $\rightarrow$ 81.35} & \textcolor[rgb]{0.0,0,0}{61.92 $\rightarrow$ 55.69}
\\
ATAE-LSTM \cite{wang2016attention}& \textcolor[rgb]{0.0,0,0}{85.60 $\rightarrow$ 28.90} & \textcolor[rgb]{0.0,0,0}{67.02 $\rightarrow$ 23.84} & \textcolor[rgb]{0.0,0,0}{86.60 $\rightarrow$ 60.74} & \textcolor[rgb]{0.0,0,0}{65.41 $\rightarrow$ 41.46} & \textcolor[rgb]{0.0,0,0}{85.60 $\rightarrow$ 44.39} & \textcolor[rgb]{0.0,0,0}{67.02 $\rightarrow$ 36.40}
\\
MemNet \cite{tang2016aspect}& \textcolor[rgb]{0.0,0,0}{81.46 $\rightarrow$ 19.30} & \textcolor[rgb]{0.0,0,0}{54.57 $\rightarrow$ 17.77} & \textcolor[rgb]{0.0,0,0}{83.68 $\rightarrow$ 72.95} & \textcolor[rgb]{0.0,0,0}{55.39 $\rightarrow$ 45.14} & \textcolor[rgb]{0.0,0,0}{81.46 $\rightarrow$ 63.62} & \textcolor[rgb]{0.0,0,0}{54.57 $\rightarrow$ 39.36}
\\
IAN \cite{ma2017interactive}& \textcolor[rgb]{0.0,0,0}{83.83 $\rightarrow$ 17.71} & \textcolor[rgb]{0.0,0,0}{58.91 $\rightarrow$ 18.12} & \textcolor[rgb]{0.0,0,0}{84.88 $\rightarrow$ 73.06} & \textcolor[rgb]{0.0,0,0}{56.91 $\rightarrow$ 45.87} & \textcolor[rgb]{0.0,0,0}{83.83 $\rightarrow$ 56.61} & \textcolor[rgb]{0.0,0,0}{58.91 $\rightarrow$ 37.08}
\\
TNet \cite{li2018transformation}& \textcolor[rgb]{0.0,0,0}{87.37 $\rightarrow$ 24.58} & \textcolor[rgb]{0.0,0,0}{66.29 $\rightarrow$ 25.00} & \textcolor[rgb]{0.0,0,0}{87.86 $\rightarrow$ 75.00} & \textcolor[rgb]{0.0,0,0}{66.15 $\rightarrow$ 49.09} & \textcolor[rgb]{0.0,0,0}{87.37 $\rightarrow$ 80.56} & \textcolor[rgb]{0.0,0,0}{66.29 $\rightarrow$ 59.68}
\\
MGAN \cite{fan2018multi}& \textcolor[rgb]{0.0,0,0}{88.15 $\rightarrow$ 26.10} & \textcolor[rgb]{0.0,0,0}{69.98 $\rightarrow$ 23.65} & \textcolor[rgb]{0.0,0,0}{89.06 $\rightarrow$ 71.95} & \textcolor[rgb]{0.0,0,0}{68.90 $\rightarrow$ 50.24} & \textcolor[rgb]{0.0,0,0}{88.15 $\rightarrow$ 70.21} & \textcolor[rgb]{0.0,0,0}{69.98 $\rightarrow$ 51.71}
\\
BERT-base \cite{devlin2019bert}& \textcolor[rgb]{0.0,0,0}{90.44  $\rightarrow$  37.17} & \textcolor[rgb]{0.0,0,0}{70.66  $\rightarrow$  30.38} & \textcolor[rgb]{0.0,0,0}{90.55  $\rightarrow$  52.46} & \textcolor[rgb]{0.0,0,0}{71.45  $\rightarrow$  32.47} & \textcolor[rgb]{0.0,0,0}{90.44  $\rightarrow$  55.96} & \textcolor[rgb]{0.0,0,0}{70.66  $\rightarrow$  37.00}
\\
BERT+aspect \cite{devlin2019bert} & \textcolor[rgb]{0.0,0,0}{90.32 $\rightarrow$ 62.59} & \textcolor[rgb]{0.0,0,0}{76.91 $\rightarrow$ 44.83} & \textcolor[rgb]{0.0,0,0}{91.41 $\rightarrow$ 57.04} & \textcolor[rgb]{0.0,0,0}{77.53 $\rightarrow$ 44.43} & \textcolor[rgb]{0.0,0,0}{90.32 $\rightarrow$ 81.58} & \textcolor[rgb]{0.0,0,0}{76.91 $\rightarrow$ 71.01}
\\
LCF-BERT \cite{zeng2019lcf}& \textcolor[rgb]{0.0,0,0}{90.32 $\rightarrow$ 53.48} & \textcolor[rgb]{0.0,0,0}{76.56 $\rightarrow$ 39.52} & \textcolor[rgb]{0.0,0,0}{90.55 $\rightarrow$ 61.09} & \textcolor[rgb]{0.0,0,0}{75.18 $\rightarrow$ 44.87} & \textcolor[rgb]{0.0,0,0}{90.32 $\rightarrow$ 86.78} & \textcolor[rgb]{0.0,0,0}{76.56 $\rightarrow$ 73.71}
\\
\textbf{Average} & \textcolor[rgb]{0.0,0,0}{86.83 $\rightarrow$ 31.16} & \textcolor[rgb]{0.0,0,0}{65.86 $\rightarrow$ 26.63} & \textcolor[rgb]{0.0,0,0}{87.78 $\rightarrow$ 67.73} & \textcolor[rgb]{0.0,0,0}{64.96 $\rightarrow$ 45.15} & \textcolor[rgb]{0.0,0,0}{86.83 $\rightarrow$ 66.55} & \textcolor[rgb]{0.0,0,0}{65.86 $\rightarrow$ 49.49}
\\
\bottomrule
\end{tabular}
}
\label{tab:task_absa}
\end{table*}

{\bf Named Entity Recognition (NER)} is a fundamental NLP task that involves determining entity boundaries and recognizing categories of named entities, which is often formalized as a sequence labeling task. To perform robustness evaluation, we choose three widely used NER datasets, including CoNLL 2003 \cite{sang2003introduction}, ACE 2005\footnote{https://catalog.ldc.upenn.edu/LDC2006T06}, and OntoNotes \cite{weischedel2012ontonotes}\footnote{https://catalog.ldc.upenn.edu/LDC2013T19}. We test 10 models under five different transformation settings using the metric of F1 score.

The changes of model performances are listed in Table \ref{tab:task_ner},  where we can observe that model performance is not noticeably influenced by \textbf{\emph{ConcatSent}}, which indicates that general transformations such as simple concatenation might have difficulty finding core defects for specific tasks. On the other hand, task-specific transformations, e.g., \textbf{\emph{CrossCategory}} and \textbf{\emph{SwapLonger}}, induce a significant performance drop of all tested systems. It indicates that most existing NER models are inadequate to deal with inherent challenges of NER, such as the problem of combinatorial ambiguity and OOV entities. 

\begin{table*}[!t]
\small
\centering
\caption{F1 score on the CoNLL 2003 dataset.
}
\resizebox{\textwidth}{!}{
\begin{tabular}{lccccc}
\toprule
\multirow{2}{*}{Model}
& \multicolumn{1}{c}{\textbf{\emph{ConcatSent}}}
& \multicolumn{1}{c}{\textbf{\emph{CrossCategory}}}
& \multicolumn{1}{c}{\textbf{\emph{EntTypos}}}
& \multicolumn{1}{c}{\textbf{\emph{OOV}}}
& \multicolumn{1}{c}{\textbf{\emph{SwapLonger}}}
\\
&  Ori. $\rightarrow$ Trans.
&  Ori. $\rightarrow$ Trans.
&  Ori. $\rightarrow$ Trans.
&  Ori. $\rightarrow$ Trans.
&  Ori. $\rightarrow$ Trans.
\\
\midrule
\multicolumn{6}{l}{\textbf{\textit{CoNLL 2003}}} \\
CNN-LSTM-CRF \cite{ma2016end}& \textcolor[rgb]{0.0,0,0}{90.61 $\rightarrow$ 87.99} & \textcolor[rgb]{0.0,0,0}{90.59 $\rightarrow$ 44.18} & \textcolor[rgb]{0.0,0,0}{91.25 $\rightarrow$ 79.10} & \textcolor[rgb]{0.0,0,0}{90.59 $\rightarrow$ 58.99} & \textcolor[rgb]{0.0,0,0}{90.59 $\rightarrow$ 61.15}
\\
LSTM-CRF \cite{lample2016neural}& \textcolor[rgb]{0.0,0,0}{88.49 $\rightarrow$ 86.88} & \textcolor[rgb]{0.0,0,0}{88.48 $\rightarrow$ 41.33} & \textcolor[rgb]{0.0,0,0}{89.31 $\rightarrow$ 74.32} & \textcolor[rgb]{0.0,0,0}{88.48 $\rightarrow$ 43.55} & \textcolor[rgb]{0.0,0,0}{88.48 $\rightarrow$ 54.50}
\\
LM-LSTM-CRF \cite{liu2018empower}& \textcolor[rgb]{0.0,0,0}{90.89 $\rightarrow$ 88.21} & \textcolor[rgb]{0.0,0,0}{90.88 $\rightarrow$ 44.28} & \textcolor[rgb]{0.0,0,0}{91.54 $\rightarrow$ 82.90} & \textcolor[rgb]{0.0,0,0}{90.88 $\rightarrow$ 70.40} & \textcolor[rgb]{0.0,0,0}{90.88 $\rightarrow$ 65.43} 
\\
Elmo \cite{peters2018deep}& \textcolor[rgb]{0.0,0,0}{91.80 $\rightarrow$ 90.67} & \textcolor[rgb]{0.0,0,0}{91.79 $\rightarrow$ 44.13} & \textcolor[rgb]{0.0,0,0}{92.48 $\rightarrow$ 86.19} & \textcolor[rgb]{0.0,0,0}{91.79 $\rightarrow$ 68.10} & \textcolor[rgb]{0.0,0,0}{91.79 $\rightarrow$ 61.82} 
\\
Flair \cite{akbik2018contextual}& \textcolor[rgb]{0.0,0,0}{92.25 $\rightarrow$ 90.73} & \textcolor[rgb]{0.0,0,0}{92.24 $\rightarrow$ 45.30} & \textcolor[rgb]{0.0,0,0}{93.05 $\rightarrow$ 86.78} & \textcolor[rgb]{0.0,0,0}{92.24 $\rightarrow$ 73.45} & \textcolor[rgb]{0.0,0,0}{92.24 $\rightarrow$ 66.13} 
\\
Pooled-Flair \cite{akbik2019pooled}& \textcolor[rgb]{0.0,0,0}{91.90 $\rightarrow$ 90.45} & \textcolor[rgb]{0.0,0,0}{91.88 $\rightarrow$ 43.64} & \textcolor[rgb]{0.0,0,0}{92.72 $\rightarrow$ 86.38} & \textcolor[rgb]{0.0,0,0}{91.88 $\rightarrow$ 71.70} & \textcolor[rgb]{0.0,0,0}{91.88 $\rightarrow$ 67.92} 
\\
TENER \cite{yan2019tener}& \textcolor[rgb]{0.0,0,0}{91.36 $\rightarrow$ 90.27} & \textcolor[rgb]{0.0,0,0}{91.35 $\rightarrow$ 45.43} & \textcolor[rgb]{0.0,0,0}{92.01 $\rightarrow$ 82.26} & \textcolor[rgb]{0.0,0,0}{91.35 $\rightarrow$ 55.67} & \textcolor[rgb]{0.0,0,0}{91.35 $\rightarrow$ 51.10} 
\\
GRN \cite{chen2019grn}& \textcolor[rgb]{0.0,0,0}{91.57 $\rightarrow$ 89.30} & \textcolor[rgb]{0.0,0,0}{91.56 $\rightarrow$ 42.90} & \textcolor[rgb]{0.0,0,0}{92.29 $\rightarrow$ 82.72} & \textcolor[rgb]{0.0,0,0}{91.56 $\rightarrow$ 68.20} & \textcolor[rgb]{0.0,0,0}{91.56 $\rightarrow$ 65.38} 
\\
BERT-base (cased) \cite{devlin2019bert}& \textcolor[rgb]{0.0,0,0}{91.43 $\rightarrow$ 89.91} & \textcolor[rgb]{0.0,0,0}{91.42 $\rightarrow$ 44.42} & \textcolor[rgb]{0.0,0,0}{92.20 $\rightarrow$ 85.02} & \textcolor[rgb]{0.0,0,0}{91.42 $\rightarrow$ 68.71} & \textcolor[rgb]{0.0,0,0}{91.42 $\rightarrow$ 79.28} 
\\
BERT-base (uncased) \cite{devlin2019bert}& \textcolor[rgb]{0.0,0,0}{90.41 $\rightarrow$ 90.05} & \textcolor[rgb]{0.0,0,0}{90.40 $\rightarrow$ 47.19} & \textcolor[rgb]{0.0,0,0}{91.25 $\rightarrow$ 81.25} & \textcolor[rgb]{0.0,0,0}{90.40 $\rightarrow$ 64.46} & \textcolor[rgb]{0.0,0,0}{90.40 $\rightarrow$ 78.26} 
\\
\textbf{Average} & \textcolor[rgb]{0.0,0,0}{91.07 $\rightarrow$ 89.45} & \textcolor[rgb]{0.0,0,0}{91.06 $\rightarrow$ 44.28} & \textcolor[rgb]{0.0,0,0}{91.81 $\rightarrow$ 82.69} & \textcolor[rgb]{0.0,0,0}{91.06 $\rightarrow$ 64.32} & \textcolor[rgb]{0.0,0,0}{91.06 $\rightarrow$ 65.10}
\\
\bottomrule
\end{tabular}
}
\label{tab:task_ner}
\end{table*}

{\bf Machine Reading Comprehension (MRC)} aims to comprehend the context of given articles and answer the questions based on them. Various types of MRC datasets exist, such as cloze-style reading comprehension \cite{hermann2015teaching} and span-extraction reading comprehension \cite{rajpurkar2016squad}. In this work, we focus on the span-extraction scenario and choose two typical MRC datasets, namely, SQuAD 1.0 \cite{rajpurkar2016squad} and SQuAD 2.0 \cite{rajpurkar2018know}. Since the official test set is not publicly released, we use the development set to produce transformed samples. Following previous works \cite{seo2016bidirectional,chen2017reading}, we adopt Exact Match (EM) and F1 score as our evaluation metrics. Table \ref{tab:task_mrc} presents the results of different systems on the original and enriched development set of SQuAD 1.0 dataset.

From the table, we find that \textbf{\emph{ModifyPos}} can hardly hurt model performances, which indicates that span-extraction models are insensitive to answer positions. Meanwhile, the modification of text contents, e.g., \textbf{\emph{PerturbAnswer}}, can bring a drastic performance degradation to all systems. It reflects that models might overfit on dataset-specific features and fail to identify answer spans that are perturbed into unseen patterns even when their meanings are unchanged.

\begin{table*}[!t]
\small
\centering
\caption{Exact Match (EM) and F1 score on the SQuAD 1.0 dataset.
}
\resizebox{\textwidth}{!}{
\begin{tabular}{lcccccc}
\toprule
\multirow{2}{*}{Model}
& \multicolumn{2}{c}{\textbf{\emph{ModifyPos}} (Ori.$\rightarrow$Trans.)}
& \multicolumn{2}{c}{\textbf{\emph{AddSentDiverse}} (Ori.$\rightarrow$Trans.)}
& \multicolumn{2}{c}{\textbf{\emph{PerturbAnswer}} (Ori.$\rightarrow$Trans.)}
\\
& Exact Match & F1 Score
& Exact Match & F1 Score
& Exact Match & F1 Score
\\
\midrule
\multicolumn{7}{l}{\textbf{\textit{SQuAD 1.0}}} \\
BiDAF \cite{seo2016bidirectional}& \textcolor[rgb]{0.0,0,0}{68.93 $\rightarrow$ 68.64} & \textcolor[rgb]{0.0,0,0}{78.09 $\rightarrow$ 77.52} & \textcolor[rgb]{0.0,0,0}{68.10 $\rightarrow$ 22.68} & \textcolor[rgb]{0.0,0,0}{77.45 $\rightarrow$ 26.07} & \textcolor[rgb]{0.0,0,0}{68.27 $\rightarrow$ 51.24} & \textcolor[rgb]{0.0,0,0}{77.50 $\rightarrow$ 63.76}
\\
BiDAF$^+$ \cite{seo2016bidirectional}& \textcolor[rgb]{0.0,0,0}{69.60 $\rightarrow$ 67.58} & \textcolor[rgb]{0.0,0,0}{78.91 $\rightarrow$ 76.72} & \textcolor[rgb]{0.0,0,0}{68.88 $\rightarrow$ 22.71} & \textcolor[rgb]{0.0,0,0}{78.21 $\rightarrow$ 26.60} & \textcolor[rgb]{0.0,0,0}{68.91 $\rightarrow$ 52.19} & \textcolor[rgb]{0.0,0,0}{78.24 $\rightarrow$ 64.55}
\\
DrQA \cite{chen2017reading}& \textcolor[rgb]{0.0,0,0}{70.99 $\rightarrow$ 69.99} & \textcolor[rgb]{0.0,0,0}{80.20 $\rightarrow$ 78.67} & \textcolor[rgb]{0.0,0,0}{70.34 $\rightarrow$ 35.34} & \textcolor[rgb]{0.0,0,0}{79.62 $\rightarrow$ 40.56} & \textcolor[rgb]{0.0,0,0}{70.19 $\rightarrow$ 52.32} & \textcolor[rgb]{0.0,0,0}{79.52 $\rightarrow$ 64.85}
\\
R-Net \cite{wang2017r}& \textcolor[rgb]{0.0,0,0}{72.06 $\rightarrow$ 70.79} & \textcolor[rgb]{0.0,0,0}{80.56 $\rightarrow$ 78.96} & \textcolor[rgb]{0.0,0,0}{71.31 $\rightarrow$ 26.55} & \textcolor[rgb]{0.0,0,0}{79.83 $\rightarrow$ 30.63} & \textcolor[rgb]{0.0,0,0}{71.35 $\rightarrow$ 54.15} & \textcolor[rgb]{0.0,0,0}{79.87 $\rightarrow$ 66.13}
\\
FusionNet \cite{huang2018fusionnet}& \textcolor[rgb]{0.0,0,0}{73.00 $\rightarrow$ 71.60} & \textcolor[rgb]{0.0,0,0}{82.01 $\rightarrow$ 80.38} & \textcolor[rgb]{0.0,0,0}{72.21 $\rightarrow$ 34.40} & \textcolor[rgb]{0.0,0,0}{81.28 $\rightarrow$ 39.33} & \textcolor[rgb]{0.0,0,0}{72.47 $\rightarrow$ 54.90} & \textcolor[rgb]{0.0,0,0}{81.44 $\rightarrow$ 67.49}
\\
QANet \cite{yu2018qanet}& \textcolor[rgb]{0.0,0,0}{71.52 $\rightarrow$ 71.27} & \textcolor[rgb]{0.0,0,0}{79.98 $\rightarrow$ 79.79} & \textcolor[rgb]{0.0,0,0}{70.67 $\rightarrow$ 19.34} & \textcolor[rgb]{0.0,0,0}{79.32 $\rightarrow$ 22.09} & \textcolor[rgb]{0.0,0,0}{70.86 $\rightarrow$ 55.13} & \textcolor[rgb]{0.0,0,0}{79.45 $\rightarrow$ 67.36}
\\
BERT \cite{devlin2019bert} & \textcolor[rgb]{0.0,0,0}{79.95 $\rightarrow$ 79.81} & \textcolor[rgb]{0.0,0,0}{87.68 $\rightarrow$ 87.25} & \textcolor[rgb]{0.0,0,0}{79.25 $\rightarrow$ 27.93} & \textcolor[rgb]{0.0,0,0}{87.09 $\rightarrow$ 32.47} & \textcolor[rgb]{0.0,0,0}{79.30 $\rightarrow$ 62.48} & \textcolor[rgb]{0.0,0,0}{87.13 $\rightarrow$ 75.40}
\\
ALBERT-V2 \cite{lan2019albert}& \textcolor[rgb]{0.0,0,0}{85.31 $\rightarrow$ 84.24} & \textcolor[rgb]{0.0,0,0}{91.76 $\rightarrow$ 90.82} & \textcolor[rgb]{0.0,0,0}{84.70 $\rightarrow$ 35.87} & \textcolor[rgb]{0.0,0,0}{91.27 $\rightarrow$ 40.45} & \textcolor[rgb]{0.0,0,0}{84.63 $\rightarrow$ 68.80} & \textcolor[rgb]{0.0,0,0}{91.26 $\rightarrow$ 80.52}
\\
XLNet \cite{yang2019xlnet}& \textcolor[rgb]{0.0,0,0}{81.79 $\rightarrow$ 81.13} & \textcolor[rgb]{0.0,0,0}{89.81 $\rightarrow$ 88.94} & \textcolor[rgb]{0.0,0,0}{81.37 $\rightarrow$ 32.12} & \textcolor[rgb]{0.0,0,0}{89.50 $\rightarrow$ 37.48} & \textcolor[rgb]{0.0,0,0}{81.30 $\rightarrow$ 67.15} & \textcolor[rgb]{0.0,0,0}{89.45 $\rightarrow$ 80.15}
\\
DistillBERT \cite{sanh2019distilbert}& \textcolor[rgb]{0.0,0,0}{79.96 $\rightarrow$ 79.10} & \textcolor[rgb]{0.0,0,0}{87.56 $\rightarrow$ 86.69} & \textcolor[rgb]{0.0,0,0}{79.43 $\rightarrow$ 25.53} & \textcolor[rgb]{0.0,0,0}{87.10 $\rightarrow$ 29.60} & \textcolor[rgb]{0.0,0,0}{79.35 $\rightarrow$ 62.21} & \textcolor[rgb]{0.0,0,0}{87.04 $\rightarrow$ 74.92}
\\
\textbf{Average} & \textcolor[rgb]{0.0,0,0}{75.31 $\rightarrow$ 74.42} & \textcolor[rgb]{0.0,0,0}{83.65 $\rightarrow$ 82.57} & \textcolor[rgb]{0.0,0,0}{74.63 $\rightarrow$ 28.25} & \textcolor[rgb]{0.0,0,0}{83.07 $\rightarrow$ 32.53} & \textcolor[rgb]{0.0,0,0}{74.66 $\rightarrow$ 58.06} & \textcolor[rgb]{0.0,0,0}{83.09 $\rightarrow$ 70.51}
\\
\bottomrule
\end{tabular}
}
\label{tab:task_mrc}
\end{table*}

{\bf Natural Language Inference (NLI)}, also known as recognizing textual entailment (RTE), is the task of determining if a natural language hypothesis can be inferred from a given premise in a justifiable manner. As a benchmark task for natural language understanding, NLI has been widely studied; further, many neural-based sentence encoders, especially the pretrained models, have been shown to consistently achieve high accuracies. In order to check whether it is the semantic understanding or the pattern matching that leads to a good model performance, we conduct experiments to analyze the current mainstream pretrained sentence encoders. 

Table \ref{tab:task_nli} lists the accuracy of the eight models on the MultiNLI\footnote{We show the results of mismatched part of the dataset.}\cite{williams-etal-2018-broad} dataset. From Table \ref{tab:task_nli}, we can observe that (1) \textbf{\emph{NumWord}}, on average, induces the greatest performance drop, as it requires the model to perform numerical reasoning for correct semantic inference. (2) \textbf{\emph{SwapAnt}} makes the average performance of the models drop by up to 23.33\%. It indicates that the models cannot handle the semantic contradiction expressed by the antonyms (not explicit negation) between premise-hypothesis pairs well. (3) \textbf{\emph{AddSent}} also makes the model performance drop significantly, indicating that model's ability to filter out irrelevant information needs to be improved. (4) Our transformation strategy, especially \textbf{\emph{Overlap}}, generates many premise-hypothesis pairs with large word overlap but different semantics, which successfully confuse all the systems. (5) Improved pretrained models (e.g. XLNet) perform better than the original BERT model, which reflects that adequate pretraining corpora and suitable pretraining strategies help to improve the generalization performance of the models.

\begin{table*}[!t]
\small
\centering
\caption{Model accuracy on the MultiNLI dataset.}
\begin{tabular}{lcccc}
\toprule
\multirow{2}{*}{Model}
& \multicolumn{1}{c}{\textbf{\emph{SwapAnt}}}
& \multicolumn{1}{c}{\textbf{\emph{AddSent}}}
& \multicolumn{1}{c}{\textbf{\emph{NumWord}}}
& \multicolumn{1}{c}{\textbf{\emph{Overlap}}}
\\
&  Ori. $\rightarrow$ Trans.
&  Ori. $\rightarrow$ Trans.
&  Ori. $\rightarrow$ Trans.
&  Ori. $\rightarrow$ Trans.
\\
\midrule
\multicolumn{5}{l}{\textbf{\textit{MultiNLI}}} \\
BERT-base \cite{devlin2019bert} & \textcolor[rgb]{0.0,0,0}{85.10 $\rightarrow$ 55.69} & \textcolor[rgb]{0.0,0,0}{84.43 $\rightarrow$ 55.27} & \textcolor[rgb]{0.0,0,0}{82.97 $\rightarrow$ 49.16} & \textcolor[rgb]{0.0,0,0}{None $\rightarrow$ 62.67}
\\
BERT-large \cite{devlin2019bert} & \textcolor[rgb]{0.0,0,0}{87.84 $\rightarrow$ 61.18} & \textcolor[rgb]{0.0,0,0}{86.36 $\rightarrow$ 58.19} & \textcolor[rgb]{0.0,0,0}{85.42 $\rightarrow$ 54.19} & \textcolor[rgb]{0.0,0,0}{None $\rightarrow$ 70.65}
\\
XLNet-base\cite{yang2019xlnet} & \textcolor[rgb]{0.0,0,0}{87.45 $\rightarrow$ 70.98} & \textcolor[rgb]{0.0,0,0}{86.33 $\rightarrow$ 57.65} & \textcolor[rgb]{0.0,0,0}{85.55 $\rightarrow$ 48.77} & \textcolor[rgb]{0.0,0,0}{None $\rightarrow$ 70.35}
\\
XLNet-large\cite{yang2019xlnet} & \textcolor[rgb]{0.0,0,0}{89.41 $\rightarrow$ 75.69} & \textcolor[rgb]{0.0,0,0}{88.63 $\rightarrow$ 63.37} & \textcolor[rgb]{0.0,0,0}{86.84 $\rightarrow$ 51.35} & \textcolor[rgb]{0.0,0,0}{None $\rightarrow$ 78.09}
\\
RoBERTa-base\cite{delobelle-etal-2020-robbert} & \textcolor[rgb]{0.0,0,0}{87.45 $\rightarrow$ 63.53} & \textcolor[rgb]{0.0,0,0}{87.13 $\rightarrow$ 57.25} & \textcolor[rgb]{0.0,0,0}{86.58 $\rightarrow$ 50.32} & \textcolor[rgb]{0.0,0,0}{None $\rightarrow$ 75.49}
\\
RoBERTa-large\cite{delobelle-etal-2020-robbert} & \textcolor[rgb]{0.0,0,0}{92.16 $\rightarrow$ 74.90} & \textcolor[rgb]{0.0,0,0}{90.12 $\rightarrow$ 67.73} & \textcolor[rgb]{0.0,0,0}{88.65 $\rightarrow$ 54.71} & \textcolor[rgb]{0.0,0,0}{None $\rightarrow$ 73.14}
\\
ALBERT-base-v2 \cite{lan2019albert} & \textcolor[rgb]{0.0,0,0}{87.45 $\rightarrow$ 50.20} & \textcolor[rgb]{0.0,0,0}{84.09 $\rightarrow$ 53.59} & \textcolor[rgb]{0.0,0,0}{82.97 $\rightarrow$ 49.42} & \textcolor[rgb]{0.0,0,0}{None $\rightarrow$ 67.15}
\\
ALBERT-xxlarge-v2 \cite{lan2019albert}& \textcolor[rgb]{0.0,0,0}{91.76 $\rightarrow$ 69.80} & \textcolor[rgb]{0.0,0,0}{89.89 $\rightarrow$ 79.11} & \textcolor[rgb]{0.0,0,0}{89.03 $\rightarrow$ 46.84} & \textcolor[rgb]{0.0,0,0}{None $\rightarrow$ 74.92}
\\
\textbf{Average} & \textcolor[rgb]{0.0,0,0}{88.58 $\rightarrow$ 65.25} & \textcolor[rgb]{0.0,0,0}{87.12 $\rightarrow$ 61.52} & \textcolor[rgb]{0.0,0,0}{86.00 $\rightarrow$ 50.60} & \textcolor[rgb]{0.0,0,0}{None $\rightarrow$ 71.56}
\\
\bottomrule
\end{tabular}
\label{tab:task_nli}
\end{table*}

{\bf Chinese Word Segmentation (CWS)}, the first step in many Chinese information processing systems, aims to segment Chinese sentences into word lists. Ambiguity and out-of-vocabulary (OOV) words are two main obstacles to be solved in this task. We conduct experiments to analyze the robustness of word segmentation models in the face of difficult examples such as ambiguous and OOV words. 

Table \ref{tab:task_cws} shows the F1 score of eight different CWS models on the CTB6 dataset \cite{Xia}: It is obvious that all the CWS models achieve a high F1 score. However, \textbf{\emph{SwapName}} generate words with intersection ambiguity by modifying the last name in the text, which reduces the model score by an average of 3.16\%. \textbf{\emph{SwapNum}}, \textbf{\emph{SwapContraction}} generate long quantifiers and proper nouns, which drops the average F1 score of the models drop by up to 4.88\% and 5.83\%, respectively. These long words may contain combinatorial ambiguity and OOV words. \textbf{\emph{SwapSyn}} generate synonyms for words in the original text, which may also introduce OOV words and cause model performance degradation.

\begin{table*}[!t]
\small
\centering
\caption{F1 score on the CTB$6$ dataset.}
\resizebox{\textwidth}{!}{
\begin{tabular}{lccccc}
\toprule
\multirow{2}{*}{Model}
& \multicolumn{1}{c}{\textbf{\emph{SwapName}}}
& \multicolumn{1}{c}{\textbf{\emph{SwapNum}}}
& \multicolumn{1}{c}{\textbf{\emph{SwapVerb}}}
& \multicolumn{1}{c}{\textbf{\emph{SwapContraction}}}
& \multicolumn{1}{c}{\textbf{\emph{SwapSyn}}}
\\
&  Ori. $\rightarrow$ Trans.
&  Ori. $\rightarrow$ Trans.
&  Ori. $\rightarrow$ Trans.
&  Ori. $\rightarrow$ Trans.
&  Ori. $\rightarrow$ Trans.
\\
\midrule
\multicolumn{5}{l}{\textbf{\textit{CTB6}}} \\
FMM\footnotemark[1] & \textcolor[rgb]{0.0,0,0}{82.13 $\rightarrow$ 78.39} & \textcolor[rgb]{0.0,0,0}{83.62 $\rightarrow$ 79.88} & \textcolor[rgb]{0.0,0,0}{82.03 $\rightarrow$ 78.14} &
\textcolor[rgb]{0.0,0,0}{84.25 $\rightarrow$ 79.11} &
\textcolor[rgb]{0.0,0,0}{83.97 $\rightarrow$ 79.26} 
\\
BMM\footnotemark[1] & \textcolor[rgb]{0.0,0,0}{83.21 $\rightarrow$ 79.28} & \textcolor[rgb]{0.0,0,0}{83.91 $\rightarrow$ 80.11} & \textcolor[rgb]{0.0,0,0}{82.45 $\rightarrow$ 78.61} & 
\textcolor[rgb]{0.0,0,0}{84.82 $\rightarrow$ 79.51} &
\textcolor[rgb]{0.0,0,0}{84.41 $\rightarrow$ 79.75} 
\\
CRF\footnotemark[2] & \textcolor[rgb]{0.0,0,0}{93.80 $\rightarrow$ 91.70} & \textcolor[rgb]{0.0,0,0}{93.30 $\rightarrow$ 89.33} & \textcolor[rgb]{0.0,0,0}{91.13 $\rightarrow$ 87.32} & 
\textcolor[rgb]{0.0,0,0}{94.20 $\rightarrow$ 87.83} &
\textcolor[rgb]{0.0,0,0}{93.50 $\rightarrow$ 92.00} 
\\
CWS-LSTM \cite{chen-etal-2015-long}& \textcolor[rgb]{0.0,0,0}{94.87 $\rightarrow$ 91.56} & \textcolor[rgb]{0.0,0,0}{95.25 $\rightarrow$ 91.32} & \textcolor[rgb]{0.0,0,0}{93.16 $\rightarrow$ 88.91} & 
\textcolor[rgb]{0.0,0,0}{95.47 $\rightarrow$ 88.88} &
\textcolor[rgb]{0.0,0,0}{94.84 $\rightarrow$ 93.01} 
\\
CWS \cite{cai-zhao-2016-neural}& \textcolor[rgb]{0.0,0,0}{94.96 $\rightarrow$ 91.31} & \textcolor[rgb]{0.0,0,0}{94.12 $\rightarrow$ 86.42} & \textcolor[rgb]{0.0,0,0}{92.42 $\rightarrow$ 87.92} & 
\textcolor[rgb]{0.0,0,0}{94.91 $\rightarrow$ 91.02} &
\textcolor[rgb]{0.0,0,0}{94.02 $\rightarrow$ 92.85} 
\\
GreedyCWS \cite{cai-etal-2017-fast}& \textcolor[rgb]{0.0,0,0}{95.18 $\rightarrow$ 91.74} & \textcolor[rgb]{0.0,0,0}{94.04 $\rightarrow$ 86.75} & \textcolor[rgb]{0.0,0,0}{93.27 $\rightarrow$ 88.54} & 
\textcolor[rgb]{0.0,0,0}{94.83 $\rightarrow$ 88.58} &
\textcolor[rgb]{0.0,0,0}{94.61 $\rightarrow$ 93.07} 
\\
Sub-CWS \cite{yang-etal-2019-subword}& \textcolor[rgb]{0.0,0,0}{95.72 $\rightarrow$ 92.92} & \textcolor[rgb]{0.0,0,0}{96.92 $\rightarrow$ 92.26} & \textcolor[rgb]{0.0,0,0}{94.01 $\rightarrow$ 89.26} & 
\textcolor[rgb]{0.0,0,0}{96.51 $\rightarrow$ 89.49} &
\textcolor[rgb]{0.0,0,0}{96.15 $\rightarrow$ 94.75} 
\\
MCCWS \cite{qiu-etal-2020-concise} & \textcolor[rgb]{0.0,0,0}{92.30 $\rightarrow$ 89.97} & \textcolor[rgb]{0.0,0,0}{92.85 $\rightarrow$ 88.94} & \textcolor[rgb]{0.0,0,0}{89.60 $\rightarrow$ 85.76} & 
\textcolor[rgb]{0.0,0,0}{93.12 $\rightarrow$ 87.03} &
\textcolor[rgb]{0.0,0,0}{92.36 $\rightarrow$ 89.77} 
\\
\textbf{Average} & \textcolor[rgb]{0.0,0,0}{91.52 $\rightarrow$ 88.36} & \textcolor[rgb]{0.0,0,0}{91.75 $\rightarrow$ 86.88} & \textcolor[rgb]{0.0,0,0}{89.76 $\rightarrow$ 85.56} & 
\textcolor[rgb]{0.0,0,0}{92.26 $\rightarrow$ 86.43} &
\textcolor[rgb]{0.0,0,0}{91.73 $\rightarrow$ 89.31} 
\\
\bottomrule
\end{tabular}
}
\label{tab:task_cws}
\end{table*}
\footnotetext[1]{https://github.com/minixalpha/PyCWS}
\footnotetext[2]{https://github.com/wellecks/cws}

\subsection{Variations of Universal Transformation}

In this section, we explore the influence of universal transformations (UT) on different natural language processing (NLP) tasks. The UT strategies cover various scenarios and applicable to multiple languages and tasks, aiming to evaluate its robustness via linguistically based attacks, model bias, and dataset subpopulations. In addition, we carry out experiments to test models under different UT combinations and task-specific transformations, thereby analyzing a correlation and synergy of these strategies.

\subsubsection{Multi-Granularity Transformation}\label{multigran}

\begin{figure}[t]
\centering
  \includegraphics[width=6.3in]{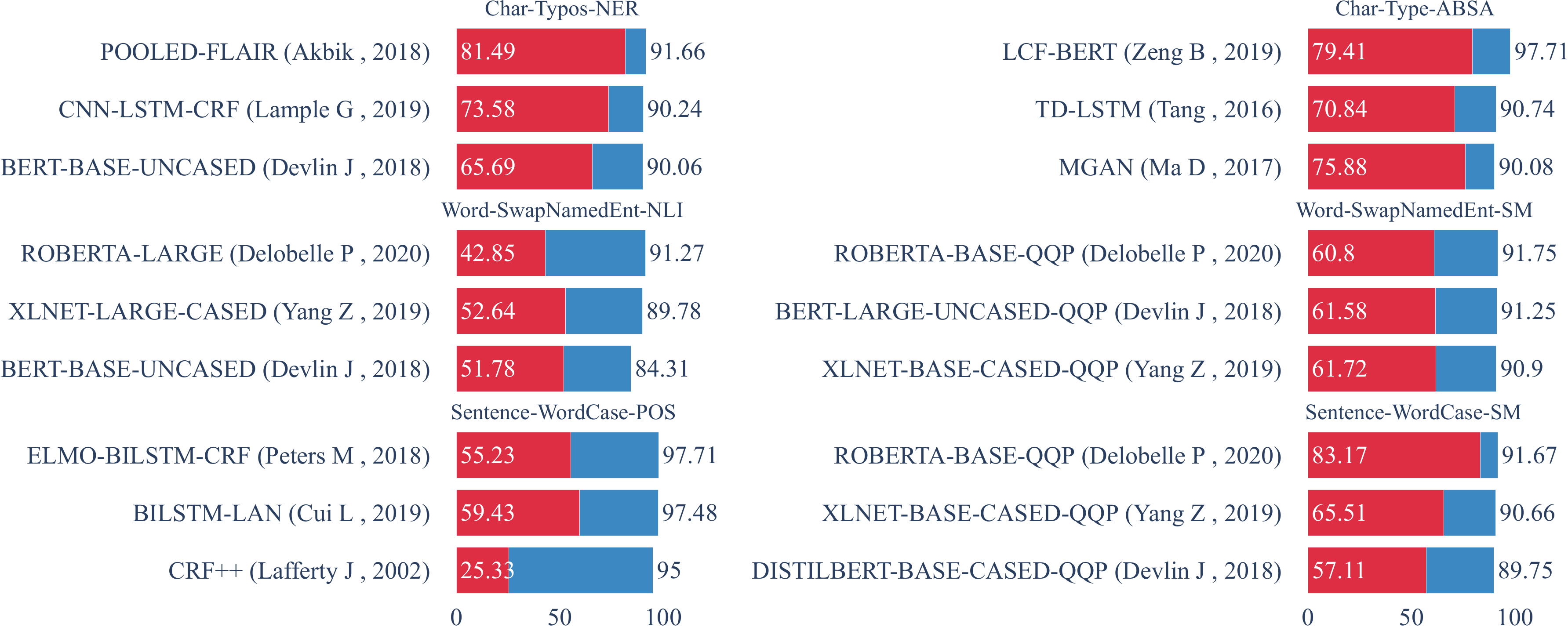}
  \caption{Accuracy results of multi-granularity universal transformations (UT). We choose \textbf{Typos}, \textbf{SwapNamedEnt}, and \textbf{WordCase} for character-level, word-level, and sentence-level UT, respectively.
  } \label{fig:ut_results}
\end{figure}

The UT strategies are guided linguistically and categorized into three levels: characters, words, and sentences. To demonstrate the influence of multi-granularity transformations on different tasks and models, we report their evaluating results under the same UT strategy and compare the original model performances with those tested from transformed samples. 

We design 3 UTs character-level, 12 UTs word-level, and 4 UTs sentence-level to explore the influence of linguistically based text transformation. Figure \ref{fig:ut_results} shows the results of multi-granularity transformed texts for several typical NLP tasks, evaluated with the {\em Accuracy} metric. We demonstrate results of \textbf{Typos}, \textbf{SwapNamedEnt}, and \textbf{WordCase} as the exemplar UTs on three levels. For \textbf{Typos}, we test models from the NER and ABSA tasks. The results show that slight changes of characters, e.g., replacement, deletion, and insertion, can drastically reduce the performance. The outcome reflects that most NLP systems are sensitive to fine-grained perturbations and vulnerable to small and precise attacks since typos may become OOV words that make the entire sentence unrecognizable. In terms of \textbf{SwapNamedEnt}, where entity words are replaced by other entities in the same category without changing the Part-Of-Speech (POS) tag and sentence structure, system performances are negatively affected by the NLI and Semantic Matching (SM) tasks. Different from \textbf{Typos}, entity replacement does not always create OOV words, as the new entity might also appear in the training data. However, the NLP systems tend to learn underlying patterns and enhance co-occurrence relationships of words rather than logic and facts, difficult to apply to other rare samples. For \textbf{WordCase}, we convert all characters into uppercase letters without changing sentence structures. In the POS-Tagging and SM task, almost all evaluated systems show a significant performance drop, indicating that they cannot deal with cased texts. This issue is difficult to be ignored because cased texts are based on important information of emphasis, emotion, and special entities.

\subsubsection{Gender and Location Bias} 

\begin{figure}[t]
\centering
  \includegraphics[width=6.3in]{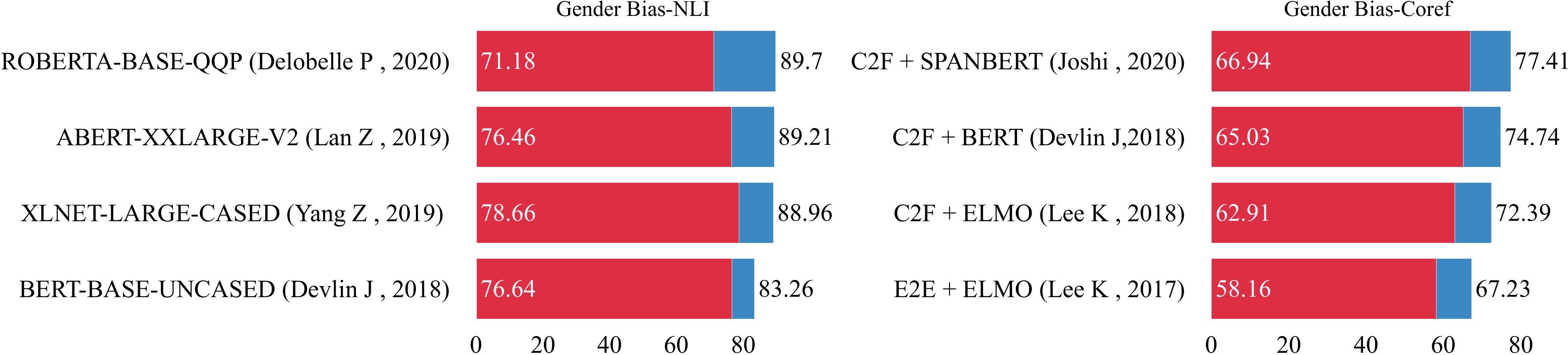}
  \caption{Results of gender bias transformations. We replace human names by female names and perform robustness evaluation in NLI and Coref tasks.
  } \label{fig:bias_results}
\end{figure}

Gender and location bias is the  preference or prejudice toward a certain gender or location \cite{moss2012science}, exhibited in multiple parts of a NLP system, including datasets, resources, and algorithms \cite{sun2019mitigating}. Systems with gender or location biases often learn and amplify negative associations about protected groups and sensitive areas in training data \cite{zhao2017men}. In addition, they produce biased predictions and uncontrollably affect downstream applications. To analyze the underlying effects of such biases in NLP systems, we design and carry out a group of universal transformations based on the gender and location bias to evaluate their robustness on a wide range of tasks and models.

We devise an entity-based transformation for bias robustness evaluation, which detects entities of human names and locations in texts, and replaces them with human names based on gender or locations in a specific region. Figure \ref{fig:bias_results} compares the results of different systems on the biased texts. We present results of gender bias on NLI tasks and Coreference Resolution (Coref.), two representative tasks for semantic understanding. From Figure \ref{fig:bias_results}, we observe that after replacing original names with female names, all systems in NLI and Coref. suffered a serious performance drop (approximately 10\% drop in accuracy). A possible explanation is that female names are inaccessible in the training set, and the model fails to recognize unfamiliar names to achieve an accurate prediction. Accordingly, we assume that if training resources exhibit gender preference or prejudice, extra negative associations between names and labels lead to a worse situation, especially in an application that focuses on social connections.

\subsubsection{Subpopulations}

With an increase in computational power and complexity of the deep learning model, the data size for model training is increasing. The performance of a complex model varies in a different subpopulation of a large, diverse dataset. In other words, good performance in one subpopulation does not imply good performance in other subpopulation \cite{pmlr-v81-buolamwini18a}, which is of additional interest to financial and medical communities  \cite{Jagielski2020SubpopulationDP}. 
We design a group of subpopulation transformation to evaluate the underlying effects on diverse subpopulations. We perform a robustness evaluation on different NLP tasks with their representative models. We take \textit{Natural language inference} task as a case study. The experiments are carried on \textit{MultiNLI}\footnote{We use mismatched testset.} dataset using accuracy as the metric.

From table \ref{tab:task_nli_pop}, we observe that: (1) \textbf{LMSubpopulation} (i.e. the fluency of language) has no significant effect on semantic matching. (2) The semantic implication between long sentences is more difficult to deal with, as it requires the model to encode more complex context semantics. (3) Compared with questions, the model can deal with negation better. (4) Surprisingly, the NLI model can process the text pair that contains the pronouns for women better than that of men.

\begin{table*}[!t]
\small
\centering
\caption{Model accuracy on different subpopulations of MultiNLI dataset. \textbf{LMSubpopulation} generates two slices based on top 20\%(more fluent) and bottom 20\%(less fluent) perplexity percentiles. Similarly, \textbf{LengthSubpopulation} generates two slices based on top 20\%(shorter) and bottom 20\%(longer) length percentiles. }
\resizebox{\textwidth}{20mm}{
\begin{tabular}{lcccccccc}
\toprule
\multirow{2}{*}{Model}
& \multicolumn{2}{c}{\textbf{LMSubpopulation}}
& \multicolumn{2}{c}{\textbf{LengthSubpopulation}}
& \multicolumn{2}{c}{\textbf{PhraseSubpopulation}}
& \multicolumn{2}{c}{\textbf{PrejudiceSubpopulation}}
\\
& \multicolumn{1}{c}{(0\%-20\%)}
& \multicolumn{1}{c}{(80\%-100\%)}
& \multicolumn{1}{c}{(0\%-20\%)}
& \multicolumn{1}{c}{(80\%-100\%)}
& \multicolumn{1}{c}{(negation)}
& \multicolumn{1}{c}{(question)}
& \multicolumn{1}{c}{(man)}
& \multicolumn{1}{c}{(woman)}
\\
\midrule
\multicolumn{5}{l}{\textbf{\textit{MultiNLI}}} \\
BERT-base \cite{devlin2019bert} & \textcolor[rgb]{0.0,0,0}{84.94} & \textcolor[rgb]{0.0,0,0}{83.17} & \textcolor[rgb]{0.0,0,0}{86.11} & \textcolor[rgb]{0.0,0,0}{82.51} & \textcolor[rgb]{0.0,0,0}{85.62} & \textcolor[rgb]{0.0,0,0}{82.14} & \textcolor[rgb]{0.0,0,0}{83.00} & \textcolor[rgb]{0.0,0,0}{86.21}
\\
BERT-large \cite{devlin2019bert} & \textcolor[rgb]{0.0,0,0}{86.27} & \textcolor[rgb]{0.0,0,0}{85.05} & \textcolor[rgb]{0.0,0,0}{87.44} & \textcolor[rgb]{0.0,0,0}{84.65} & \textcolor[rgb]{0.0,0,0}{87.63} & \textcolor[rgb]{0.0,0,0}{85.29} & \textcolor[rgb]{0.0,0,0}{85.05} & \textcolor[rgb]{0.0,0,0}{88.60}
\\
XLNet-base\cite{yang2019xlnet} & \textcolor[rgb]{0.0,0,0}{86.11} & \textcolor[rgb]{0.0,0,0}{85.41} & \textcolor[rgb]{0.0,0,0}{87.28} & \textcolor[rgb]{0.0,0,0}{87.45} & \textcolor[rgb]{0.0,0,0}{87.77} & \textcolor[rgb]{0.0,0,0}{85.02} & \textcolor[rgb]{0.0,0,0}{84.64} & \textcolor[rgb]{0.0,0,0}{88.76}
\\
XLNet-large\cite{yang2019xlnet} & \textcolor[rgb]{0.0,0,0}{87.54} & \textcolor[rgb]{0.0,0,0}{88.05} & \textcolor[rgb]{0.0,0,0}{89.27} & \textcolor[rgb]{0.0,0,0}{86.53} & \textcolor[rgb]{0.0,0,0}{90.04} & \textcolor[rgb]{0.0,0,0}{86.38} & \textcolor[rgb]{0.0,0,0}{87.21} & \textcolor[rgb]{0.0,0,0}{88.97}
\\
RoBERTa-base\cite{delobelle-etal-2020-robbert} & \textcolor[rgb]{0.0,0,0}{86.83} & \textcolor[rgb]{0.0,0,0}{86.83} & \textcolor[rgb]{0.0,0,0}{88.61} & \textcolor[rgb]{0.0,0,0}{84.90} & \textcolor[rgb]{0.0,0,0}{87.68} & \textcolor[rgb]{0.0,0,0}{85.08} & \textcolor[rgb]{0.0,0,0}{85.92} & \textcolor[rgb]{0.0,0,0}{88.24}
\\
RoBERTa-large\cite{delobelle-etal-2020-robbert} & \textcolor[rgb]{0.0,0,0}{89.73} & \textcolor[rgb]{0.0,0,0}{89.32} & \textcolor[rgb]{0.0,0,0}{90.54} & \textcolor[rgb]{0.0,0,0}{88.82} & \textcolor[rgb]{0.0,0,0}{91.65} & \textcolor[rgb]{0.0,0,0}{88.18} & \textcolor[rgb]{0.0,0,0}{89.43} & \textcolor[rgb]{0.0,0,0}{89.61}
\\
ALBERT-base-v2 \cite{lan2019albert} & \textcolor[rgb]{0.0,0,0}{84.32} & \textcolor[rgb]{0.0,0,0}{84.29} & \textcolor[rgb]{0.0,0,0}{85.91} & \textcolor[rgb]{0.0,0,0}{81.19} & \textcolor[rgb]{0.0,0,0}{85.49} & \textcolor[rgb]{0.0,0,0}{82.24} & \textcolor[rgb]{0.0,0,0}{83.12} & \textcolor[rgb]{0.0,0,0}{85.39}
\\
ALBERT-xxlarge-v2 \cite{lan2019albert} & \textcolor[rgb]{0.0,0,0}{89.32} & \textcolor[rgb]{0.0,0,0}{88.00} & \textcolor[rgb]{0.0,0,0}{89.98} & \textcolor[rgb]{0.0,0,0}{88.97} & \textcolor[rgb]{0.0,0,0}{90.76} & \textcolor[rgb]{0.0,0,0}{88.51} & \textcolor[rgb]{0.0,0,0}{89.31} & \textcolor[rgb]{0.0,0,0}{89.80}
\\
\textbf{Average} & \textcolor[rgb]{0.0,0,0}{86.87} & \textcolor[rgb]{0.0,0,0}{86.27} & \textcolor[rgb]{0.0,0,0}{88.14} & \textcolor[rgb]{0.0,0,0}{85.29} & \textcolor[rgb]{0.0,0,0}{88.33} & \textcolor[rgb]{0.0,0,0}{85.36} & \textcolor[rgb]{0.0,0,0}{85.96} & \textcolor[rgb]{0.0,0,0}{87.95}
\\
\bottomrule
\end{tabular}}
\label{tab:task_nli_pop}
\end{table*}

\subsubsection{Combination of Transformations}
To identify model shortcomings and help participants to revise the models in the real-world development cycle, we carry out a comprehensive robustness analysis regarding the evaluation of model failure. Besides, we carry out 60 task-specific transformations to find core defects of different tasks, as described in section \ref{sec:task}, \textsf{TextFlint} offers 20 universal transformations and thousands of combinations for the generalization capabilities probe and customized analysis, respectively. Based on the different transformation and their combination, large-scale evaluations are carried out on 12 NLP tasks using the corresponding mainstream model. We demonstrate two classic tasks, \textit{Named Entity Recognition} and \textit{Natural Language Inference}, as case studies. For each task, we present results of different models under one task-specific transformation, one universal transformation, and their combination. The experimental results are displayed in table \ref{tab:task_ut_ner}.

For Named Entity Recognition task, we use \textbf{\emph{OOV}} and \textbf{SpellingError} as task-specific and general transformation, respectively. We observe some important phenomena after comparing the performance degradation caused by \textbf{\emph{OOV}}, \textbf{SpellingError}, and their combination. Although \textbf{\emph{OOV}} and \textbf{SpellingError} drop model performance significantly, the $F_1$ score of each model is more reduced on \textbf{\emph{OOV}} + \textbf{SpellingError} than either of \textbf{\emph{OOV}} and \textbf{SpellingError}. Take TENER model as an example, the performance drop in combinational transformation is 45.56, higher than the sum of the performance drop of the two-separate transformation (20.18 and 11.03, specifically).

For NLI task, we use \textbf{\emph{NumWord}} and \textbf{SwapSyn} as task-specific and universal transformation, respectively. Similarly, we observe that \textbf{\emph{NumWord}}, \textbf{SwapSyn}, and their combination drop the model accuracy by average of 35.41, 6.49, and 37.73, respectively. The outcomes indicate that the combination of several different transformation strategies makes a more challenging probe, essential for comprehensively detecting model defects.

\begin{table*}[!t]
\small
\centering
\caption{Analysis of model performance drop under \textsc{Task Specific Transformation}, \textsc{Universal Transformation} and their combination, with a focus on NER and NLI as a case study. $F_1$ score and accuracy are used a s the metric for NER and NLI, respectively.}
\resizebox{\textwidth}{!}{
\begin{tabular}{lccc}
\toprule
\multirow{2}{*}{Model}
& \multicolumn{1}{c}{\textsc{\makecell[c]{Task Specific\\Transformation}}}
& \multicolumn{1}{c}{\textsc{\makecell[c]{Universal\\Transformation}}}
& \multicolumn{1}{c}{\textsc{\makecell[c]{Combinational\\Transformation}}}
\\
&  Ori. $\rightarrow$ Trans.
&  Ori. $\rightarrow$ Trans.
&  Ori. $\rightarrow$ Trans.
\\
\midrule
\multicolumn{1}{l}{\textbf{\textit{NER(CoNLL 2003)}}} & \multicolumn{1}{c}{\textbf{\emph{OOV}}} & \multicolumn{1}{c}{\textbf{SpellingError}} & \multicolumn{1}{c}{\textbf{\emph{OOV}+SpellingError}} \\
CNN-LSTM-CRF \cite{ma2016end}& \textcolor[rgb]{0.0,0,0}{90.59 $\rightarrow$ 58.99} & \textcolor[rgb]{0.0,0,0}{90.61 $\rightarrow$ 75.89} & \textcolor[rgb]{0.0,0,0}{90.37 $\rightarrow$ 53.72}
\\
LSTM-CRF \cite{lample2016neural}& \textcolor[rgb]{0.0,0,0}{88.48 $\rightarrow$ 43.55} & \textcolor[rgb]{0.0,0,0}{88.51 $\rightarrow$ 71.53} & \textcolor[rgb]{0.0,0,0}{82.19 $\rightarrow$ 39.33}
\\
LM-LSTM-CRF \cite{liu2018empower}& \textcolor[rgb]{0.0,0,0}{90.88 $\rightarrow$ 70.40} & \textcolor[rgb]{0.0,0,0}{90.93 $\rightarrow$ 77.36} & \textcolor[rgb]{0.0,0,0}{90.59 $\rightarrow$ 64.25} 
\\
Elmo \cite{peters2018deep}& \textcolor[rgb]{0.0,0,0}{91.42 $\rightarrow$ 68.71} & \textcolor[rgb]{0.0,0,0}{92.31 $\rightarrow$ 80.75} & \textcolor[rgb]{0.0,0,0}{92.40 $\rightarrow$ 70.53} 
\\
Flair \cite{akbik2018contextual}& \textcolor[rgb]{0.0,0,0}{91.56 $\rightarrow$ 68.20} & \textcolor[rgb]{0.0,0,0}{92.76 $\rightarrow$ 83.85} & \textcolor[rgb]{0.0,0,0}{92.40 $\rightarrow$ 70.53} 
\\
Pooled-Flair \cite{akbik2019pooled}& \textcolor[rgb]{0.0,0,0}{90.40 $\rightarrow$ 64.46} & \textcolor[rgb]{0.0,0,0}{92.17 $\rightarrow$ 83.34} & \textcolor[rgb]{0.0,0,0}{91.87 $\rightarrow$ 69.35} 
\\
TENER \cite{yan2019tener}& \textcolor[rgb]{0.0,0,0}{91.88 $\rightarrow$ 71.70} & \textcolor[rgb]{0.0,0,0}{91.33 $\rightarrow$ 80.30} & \textcolor[rgb]{0.0,0,0}{91.04 $\rightarrow$ 45.48} 
\\
GRN \cite{chen2019grn}& \textcolor[rgb]{0.0,0,0}{91.35 $\rightarrow$ 55.67} & \textcolor[rgb]{0.0,0,0}{91.74 $\rightarrow$ 77.83} & \textcolor[rgb]{0.0,0,0}{91.39 $\rightarrow$ 65.06} 
\\
BERT-base (cased) \cite{devlin2019bert}& \textcolor[rgb]{0.0,0,0}{91.79 $\rightarrow$ 68.10} & \textcolor[rgb]{0.0,0,0}{91.93 $\rightarrow$ 76.44} & \textcolor[rgb]{0.0,0,0}{91.58 $\rightarrow$ 62.52} 
\\
BERT-base (uncased) \cite{devlin2019bert}& \textcolor[rgb]{0.0,0,0}{92.24 $\rightarrow$ 73.45} & \textcolor[rgb]{0.0,0,0}{90.51 $\rightarrow$ 70.00} & \textcolor[rgb]{0.0,0,0}{90.12 $\rightarrow$ 50.27} 
\\
\textbf{Average} & \textcolor[rgb]{0.0,0,0}{91.06 $\rightarrow$ 64.32} & \textcolor[rgb]{0.0,0,0}{91.28 $\rightarrow$ 77.73} & \textcolor[rgb]{0.0,0,0}{90.95 $\rightarrow$ 58.71}
\\
\midrule
\multicolumn{1}{l}{\textbf{\textit{NLI(MultiNLI)}}} & \multicolumn{1}{c}{\textbf{\emph{NumWord}}} & \multicolumn{1}{c}{\textbf{SwapSyn}} & \multicolumn{1}{c}{\textbf{\emph{NumWord}+SwapSyn}}
\\
BERT-base \cite{devlin2019bert} & \textcolor[rgb]{0.0,0,0}{82.97 $\rightarrow$ 49.16} & \textcolor[rgb]{0.0,0,0}{84.45 $\rightarrow$ 77.49} & \textcolor[rgb]{0.0,0,0}{82.97 $\rightarrow$ 44.26}
\\
BERT-large \cite{devlin2019bert} & \textcolor[rgb]{0.0,0,0}{85.42 $\rightarrow$ 54.19} & \textcolor[rgb]{0.0,0,0}{86.38 $\rightarrow$ 79.17} & \textcolor[rgb]{0.0,0,0}{85.42 $\rightarrow$ 52.90}
\\
XLNet-base\cite{yang2019xlnet} & \textcolor[rgb]{0.0,0,0}{85.55 $\rightarrow$ 48.77} & \textcolor[rgb]{0.0,0,0}{86.35 $\rightarrow$ 79.64} & \textcolor[rgb]{0.0,0,0}{85.55 $\rightarrow$ 45.16}
\\
XLNet-large\cite{yang2019xlnet} & \textcolor[rgb]{0.0,0,0}{86.84 $\rightarrow$ 51.35} & \textcolor[rgb]{0.0,0,0}{88.66 $\rightarrow$ 82.34} & \textcolor[rgb]{0.0,0,0}{86.84 $\rightarrow$ 48.39}
\\
RoBERTa-base\cite{delobelle-etal-2020-robbert} & \textcolor[rgb]{0.0,0,0}{86.58 $\rightarrow$ 50.32} & \textcolor[rgb]{0.0,0,0}{87.15 $\rightarrow$ 80.59} & \textcolor[rgb]{0.0,0,0}{86.58 $\rightarrow$ 47.10}
\\
RoBERTa-large\cite{delobelle-etal-2020-robbert} & \textcolor[rgb]{0.0,0,0}{88.65 $\rightarrow$ 54.71} & \textcolor[rgb]{0.0,0,0}{90.15 $\rightarrow$ 84.76} & \textcolor[rgb]{0.0,0,0}{88.65 $\rightarrow$ 47.10}
\\
ALBERT-base-v2 \cite{lan2019albert} & \textcolor[rgb]{0.0,0,0}{82.97 $\rightarrow$ 49.42} & \textcolor[rgb]{0.0,0,0}{84.09 $\rightarrow$ 76.96} & \textcolor[rgb]{0.0,0,0}{82.97 $\rightarrow$ 45.03}
\\
ALBERT-xxlarge-v2 \cite{lan2019albert}& \textcolor[rgb]{0.0,0,0}{89.03 $\rightarrow$ 46.84} & \textcolor[rgb]{0.0,0,0}{89.83 $\rightarrow$ 84.32} & \textcolor[rgb]{0.0,0,0}{89.03 $\rightarrow$ 46.97}
\\
\textbf{Average} & \textcolor[rgb]{0.0,0,0}{86.00 $\rightarrow$ 50.60} & \textcolor[rgb]{0.0,0,0}{87.15 $\rightarrow$ 80.66} & \textcolor[rgb]{0.0,0,0}{86.00 $\rightarrow$ 48.28}
\\
\bottomrule
\end{tabular}
}
\label{tab:task_ut_ner}
\end{table*}

\section{Analysis and Patching up}



\subsection{Analysis of Different Model Frameworks}

We adopt \textsf{TextFlint} to evaluate hundreds of models of 12 tasks, covering many model frameworks and learning schemas, ranging from traditional feature-based machine learning approaches to state-of-the-art neural networks. All evaluated models and their implementations are available publicly. We reproduce the official results and evaluate them on the transformed test samples. After model implementation, dataset transformation, and batch inspection, users will get evaluation reports on all aspects, comprehensively analyzing the robustness of a system by acquiring larger test samples. From the evaluation reports, we can easily compare the model results of the original test set with those of the transformed set, spotting the main defects of the input model and identifying the model that performs the best or worst. 

From the numerous evaluations and comparisons conducted by \textsf{TextFlint}, we have a thorough view of existing NLP systems and discover some underlying patterns about model robustness. For example. in the ABSA task (see Table \ref{tab:task_absa}), methods equipped with pre-training LMs show better performances than other models on the task-specific transformations, e.g., \textbf{\emph{AddDiff}}, where the accuracy score of BERT-Aspect drops from 90.32 to 81.58. Meanwhile, LSTM suffers a serious performance degradation from 84.42 to 44.63. The outcome reflects that the pre-training mechanism benefits from the rich external resources, and offers a better generalization ability than models trained from scratch.  

\subsection{Evaluating Industrial APIs}
In addition to the cutting-edge academic model, we analyze the mainstream commercial APIs, such as \textsc{Microsoft} Text Analytics API \footnote{https://aws.amazon.com/comprehend/}, \textsc{Google} Cloud Natural Language API \footnote{https://cloud.google.com/natural-language} and \textsc{Amazon} Comprehend API \footnote{https://azure.microsoft.com/en-us/services/cognitive-services/text-analytics/} on a classic NLP task--\textit{Named Entity Recognition}. We perform an experiment on the widely used CoNLL2003 \cite{sang2003introduction} NER dataset and evaluate the robustness using the metric of $F_1$ score. Due to the inconsistency of the NER tags between different commercial APIs and the CoNLL2003 dataset, we evaluate the $F_1$ score on three types of named entities including Persons, Locations and Organizations.

Table \ref{tab:task_api} provides  the evaluation result of the three commercial APIs. From the perspective of the transformation method, \textbf{\emph{CrossCategory}} on average induces the most performance drop. In addition, \textbf{\emph{OOV}} and \textbf{\emph{SwapLonger}} cause performance drop significantly, indicating the ability of the commercial APIs in identifying the OOV entities. However, the entities with ambiguity must be further improved. In addition, \textbf{\emph{EntTypos}} has relatively little influence on the results, showing that these APIs are robust to slight spelling errors. 

In comparison, Google API and Amazon API have a low performance on both the original data and transformed data. To identify the cause of the low performance, we randomly selected 100 data samples not processed correctly and manually analyzed the cause of the error. We find that many named entities in the CoNLL2003 dataset are in other named entities. The nested named entity is recognized by Google API but fails to be labeled in the CoNLL2003 dataset. The inconsistent labeling approach makes Google API get a lower score. In addition, we find that Amazon API has high accuracy in Person recognition, but it confuses Location with Organization, which is the reason for its low score.

\begin{table*}[!t]
\small
\centering
\caption{F1 score of commercial APIs on the CoNLL 2003 dataset.
}
\begin{tabular}{lcccc}
\toprule
\multirow{2}{*}{Model}
& \multicolumn{1}{c}{\textbf{\emph{CrossCategory}}}
& \multicolumn{1}{c}{\textbf{\emph{EntTypos}}}
& \multicolumn{1}{c}{\textbf{\emph{OOV}}}
& \multicolumn{1}{c}{\textbf{\emph{SwapLonger}}}
\\
&  Ori. $\rightarrow$ Trans.
&  Ori. $\rightarrow$ Trans.
&  Ori. $\rightarrow$ Trans.
&  Ori. $\rightarrow$ Trans.
\\
\midrule
\multicolumn{5}{l}{\textbf{\textit{CoNLL 2003}}} \\
Amazon & \textcolor[rgb]{0.0,0,0}{69.68 $\rightarrow$ 33.01} & \textcolor[rgb]{0.0,0,0}{70.19 $\rightarrow$ 65.98} & \textcolor[rgb]{0.0,0,0}{69.68 $\rightarrow$ 56.27} & \textcolor[rgb]{0.0,0,0}{69.68 $\rightarrow$ 57.63}
\\
Google & \textcolor[rgb]{0.0,0,0}{59.14 $\rightarrow$ 28.30} & \textcolor[rgb]{0.0,0,0}{62.41 $\rightarrow$ 50.87} & \textcolor[rgb]{0.0,0,0}{59.14 $\rightarrow$ 48.53} & \textcolor[rgb]{0.0,0,0}{59.14 $\rightarrow$ 53.40}
\\
Microsoft & \textcolor[rgb]{0.0,0,0}{82.69 $\rightarrow$ 43.37} & \textcolor[rgb]{0.0,0,0}{83.42 $\rightarrow$ 78.47} & \textcolor[rgb]{0.0,0,0}{82.69 $\rightarrow$ 60.18} & \textcolor[rgb]{0.0,0,0}{82.69 $\rightarrow$ 52.51}
\\
\textbf{Average} & \textcolor[rgb]{0.0,0,0}{70.50 $\rightarrow$ 34.89} & \textcolor[rgb]{0.0,0,0}{72.01 $\rightarrow$ 65.11} & \textcolor[rgb]{0.0,0,0}{70.50 $\rightarrow$ 54.99} & \textcolor[rgb]{0.0,0,0}{70.50 $\rightarrow$ 54.51}
\\
\bottomrule
\end{tabular}
\label{tab:task_api}
\end{table*}

\subsection{Patching up with Augmented Data}

After users feed the target model into \textsf{TextFlint} and customize their needs, \textsf{TextFlint} produces comprehensive transformed data in diagnosing the robustness of the target model.
Through diagnosis of dozens of transformed data, the robustness evaluation results describe model performance from the lexical, syntactic, semantic levels. \textsf{TextFlint} conveys the above evaluation results to users through visualization and tabulation reports, helping users to understand the shortcomings of the target model and design potential improvements. Moreover, \textsf{TextFlint} could generate massive augment data to address the defect of the target model. \textsf{TextFlint} contributes to the entire development cycle from evaluation to enhancement.

In Section \ref{sec:task}, we tested 10 different models for task-specific transformations on ABSA and observed significant performance degradation. 
To address the disability of these models to distinguish relevant aspects from non-target aspects, \textsf{TextFlint} generated three types of transformed data for adversarial training. We show the performance of models before/after adversarial training (Trans. $\rightarrow$ Adv.) on three task-specific transformations of the Restaurant dataset in Table \ref{tab:adversarial_training_absa}. Compared with training only on the original dataset, adversarial training has significantly improved performance in the three task-specific transformations.
The high-quality augment data generated by \textsf{TextFlint} can effectively improve the shortcomings of the target model, and all of these can be easily implemented in \textsf{TextFlint}.

\begin{table*}[!t]
\small
\centering
\caption{Model accuracy and F1 score before/after adversarial training on the SemEval 2014 dataset.
}
\resizebox{\textwidth}{!}{
\begin{tabular}{lcccccc}
\toprule
\multirow{2}{*}{Model}
& \multicolumn{2}{c}{\textbf{\emph{RevTgt}} (Trans. $\rightarrow$ Adv.)}
& \multicolumn{2}{c}{\textbf{\emph{RevNon}} (Trans. $\rightarrow$ Adv.)}
& \multicolumn{2}{c}{\textbf{\emph{AddDiff}} (Trans. $\rightarrow$ Adv.)}
\\
& Accuracy & Macro-F1
& Accuracy & Macro-F1
& Accuracy & Macro-F1
\\
\midrule
\multicolumn{7}{l}{\textbf{\textit{Restaurant Dataset}}}\\
LSTM \cite{hochreiter1997long} & 
\textcolor[rgb]{0.0,0,0}{24.91 $\rightarrow$ 21.27} & \textcolor[rgb]{0.0,0,0}{21.27 $\rightarrow$ 24.38} & \textcolor[rgb]{0.0,0,0}{69.07 $\rightarrow$ 76.80} & \textcolor[rgb]{0.0,0,0}{43.41 $\rightarrow$ 52.83} & \textcolor[rgb]{0.0,0,0}{48.76 $\rightarrow$ 78.98} & \textcolor[rgb]{0.0,0,0}{27.83 $\rightarrow$ 53.49}
\\
TD-LSTM \cite{tang2016effective}& 
\textcolor[rgb]{0.0,0,0}{16.64 $\rightarrow$ 52.42 } & \textcolor[rgb]{0.0,0,0}{15.19 $\rightarrow$ 34.08} & \textcolor[rgb]{0.0,0,0}{80.07 $\rightarrow$ 75.08} & \textcolor[rgb]{0.0,0,0}{53.12 $\rightarrow$ 46.72} & \textcolor[rgb]{0.0,0,0}{77.09 $\rightarrow$ 85.12} & \textcolor[rgb]{0.0,0,0}{50.30 $\rightarrow$ 57.61}
\\
ATAE-LSTM \cite{wang2016attention}& 
\textcolor[rgb]{0.0,0,0}{19.95 $\rightarrow$ 45.93} & \textcolor[rgb]{0.0,0,0}{19.27 $\rightarrow$ 28.13} & \textcolor[rgb]{0.0,0,0}{62.02 $\rightarrow$ 51.72} & \textcolor[rgb]{0.0,0,0}{42.38 $\rightarrow$ 34.94} & \textcolor[rgb]{0.0,0,0}{53.48 $\rightarrow$ 78.04} & \textcolor[rgb]{0.0,0,0}{39.47 $\rightarrow$ 51.66}
\\
MemNet \cite{tang2016aspect}& 
\textcolor[rgb]{0.0,0,0}{18.18 $\rightarrow$ 16.76} & \textcolor[rgb]{0.0,0,0}{12.34 $\rightarrow$ 9.57} & \textcolor[rgb]{0.0,0,0}{78.87 $\rightarrow$ 78.52} & \textcolor[rgb]{0.0,0,0}{47.63 $\rightarrow$ 29.32} & \textcolor[rgb]{0.0,0,0}{65.40 $\rightarrow$ 76.27} & \textcolor[rgb]{0.0,0,0}{34.20 $\rightarrow$ 28.85}
\\
IAN \cite{ma2017interactive}& 
\textcolor[rgb]{0.0,0,0}{19.13 $\rightarrow$ 17.00} & \textcolor[rgb]{0.0,0,0}{13.44 $\rightarrow$ 11.48} & \textcolor[rgb]{0.0,0,0}{76.12 $\rightarrow$ 80.41} & \textcolor[rgb]{0.0,0,0}{46.04 $\rightarrow$ 42.23} & \textcolor[rgb]{0.0,0,0}{59.50 $\rightarrow$ 76.15} & \textcolor[rgb]{0.0,0,0}{32.04 $\rightarrow$ 29.73}
\\
TNet \cite{li2018transformation}& 
\textcolor[rgb]{0.0,0,0}{23.25 $\rightarrow$ 53.95} & \textcolor[rgb]{0.0,0,0}{14.89 $\rightarrow$ 29.85} & \textcolor[rgb]{0.0,0,0}{79.04$\rightarrow$ 64.26} & \textcolor[rgb]{0.0,0,0}{48.88 $\rightarrow$ 38.87} & \textcolor[rgb]{0.0,0,0}{83.83 $\rightarrow$ 85.01} & \textcolor[rgb]{0.0,0,0}{58.91 $\rightarrow$ 53.95}

\\
MGAN \cite{fan2018multi}& 
\textcolor[rgb]{0.0,0,0}{14.99 $\rightarrow$ 76.26} & \textcolor[rgb]{0.0,0,0}{11.53 $\rightarrow$ 28.84} & \textcolor[rgb]{0.0,0,0}{69.42 $\rightarrow$ 14.09} & \textcolor[rgb]{0.0,0,0}{29.87 $\rightarrow$ 8.23} & \textcolor[rgb]{0.0,0,0}{66.94 $\rightarrow$ 16.76} & \textcolor[rgb]{0.0,0,0}{30.10 $\rightarrow$ 9.57}
\\
BERT-base \cite{devlin2019bert}& 
\textcolor[rgb]{0.0,0,0}{38.02  $\rightarrow$  16.76} & \textcolor[rgb]{0.0,0,0}{33.65  $\rightarrow$  9.57} & \textcolor[rgb]{0.0,0,0}{53.61  $\rightarrow$  78.52} & \textcolor[rgb]{0.0,0,0}{40.55  $\rightarrow$  29.32} & \textcolor[rgb]{0.0,0,0}{59.03  $\rightarrow$  76.26} & \textcolor[rgb]{0.0,0,0}{47.23  $\rightarrow$  28.84}
\\
BERT+aspect \cite{devlin2019bert} & 
\textcolor[rgb]{0.0,0,0}{64.93 $\rightarrow$ 72.61} & \textcolor[rgb]{0.0,0,0}{45.82 $\rightarrow$ 51.71} & \textcolor[rgb]{0.0,0,0}{56.70 $\rightarrow$ 72.50} & \textcolor[rgb]{0.0,0,0}{40.99 $\rightarrow$ 51.08} & \textcolor[rgb]{0.0,0,0}{78.39 $\rightarrow$ 92.21} & \textcolor[rgb]{0.0,0,0}{63.63 $\rightarrow$ 74.66}
\\
LCF-BERT \cite{zeng2019lcf}& 
\textcolor[rgb]{0.0,0,0}{59.91 $\rightarrow$ 76.74} & \textcolor[rgb]{0.0,0,0}{44.57 $\rightarrow$ 53.74} & \textcolor[rgb]{0.0,0,0}{57.39 $\rightarrow$ 69.93} & \textcolor[rgb]{0.0,0,0}{41.11 $\rightarrow$ 49.51} & \textcolor[rgb]{0.0,0,0}{58.91 $\rightarrow$ 76.74} & \textcolor[rgb]{0.0,0,0}{44.57 $\rightarrow$ 53.74}
\\
\textbf{Average} & 
\textcolor[rgb]{0.0,0,0}{29.99 $\rightarrow$ 44.97} & \textcolor[rgb]{0.0,0,0}{14.96 $\rightarrow$ 28.13} & \textcolor[rgb]{0.0,0,0}{68.23 $\rightarrow$ 72.96} & \textcolor[rgb]{0.0,0,0}{49.99 $\rightarrow$ 38.30} & \textcolor[rgb]{0.0,0,0}{65.13 $\rightarrow$ 74.16} & \textcolor[rgb]{0.0,0,0}{42.83 $\rightarrow$ 44.21}
\\
\bottomrule
\end{tabular}
}
\label{tab:adversarial_training_absa}
\end{table*}

\section{Related Tools and Work} 
Our work is related to many existing open-source tools and works in different areas.

\paragraph{Robustness Evaluation}
Many tools include evaluation methods for robustness. NLPAug \cite{ma2019nlpaug} is an open-source library focusing on data augmentation in NLP, which includes several transformation methods that also help to evaluate robustness. Errudite \cite{wu2019errudite} supports subpopulation for error-analysing. AllenNLP Interpret \cite{wallace-etal-2019-allennlp} includes attack methods for model interpreting.  Checklist \cite{ribeiro-etal-2020-beyond} also offers pertubations for model evaluating. These tools are only applicable to small parts of robustness evaluations, while \textsf{TextFlint} supports comprehensive evaluation methods like subpopulation, adversarial attacks, transformations and so on. 

There also exist several tools concerning robustness that are similar to our work \cite{morris2020textattack,zeng2020openattack,goel2021robustness}, which also include a wide range of evaluation methods. Our work is different from these works. First, these tools only focus on general generalization evaluations and lack task-specific evaluation designs for detecting the defects for specific tasks, while \textsf{TextFlint} supports both general and task-specific evaluations. Second, these tools lack quality evaluations on generated texts or only support automatic quality constraints \cite{morris2020textattack,zeng2020openattack}, while \textsf{TextFlint} have ensured the acceptability of each transformation method with human evaluations. Additionally, these tools provide limited analysis on the robustness evaluation results, while \textsf{TextFlint} provides a standard report that can be displayed with visualization and tabulation.

\paragraph{Interpretability and Error Analysis} There also exist several works concerning model evaluation from different perspective. AllenNLP Interpret \cite{wallace-etal-2019-allennlp}, InterpreteML \cite{nori2019interpretml}, LIT \cite{nori2019interpretml}, Manifold \cite{zhang2018manifold}, AIX360 \cite{arya2019one} cares about model interpretability, trying to understand the models' behavior through different evaluation methods. CrossCheck \cite{arendt2020crosscheck}, AllenNLP Interpret \cite{wallace-etal-2019-allennlp}, Errudite \cite{wu2019errudite} and Manifold \cite{zhang2018manifold} offer visualization and cross-model comparison for error analysis. \textsf{TextFlint} is differently-motivated yet complementary with these work, which can provide comprehensive analysis on models' defects, contributing to better model understanding.

\section{Conclusion}
We introduce \textsf{TextFlint}, a unified multilingual robustness evaluation toolkit that incorporates universal text transformation, task-specific transformation, adversarial attack, subpopulation, and their combinations to provide comprehensive robustness analysis. \textsf{TextFlint} adopts the \textbf{Customize} $\Rightarrow$ \textbf{Produce} $\Rightarrow$ \textbf{Analyze} workflow to address the challenges of integrity, acceptability, and analyzability. \textsf{TextFlint} enables practitioners to evaluate their models with just a few lines of code, and then obtain complete analytical reports. We performed large-scale empirical evaluations on state-of-the-art deep learning models, classic supervised methods, and real-world systems. Almost all models showed significant performance degradation, indicating the urgency and necessity of including robustness into NLP model evaluations.

\bibliographystyle{coling}
\bibliography{coling2020}

\appendix

\end{document}